\documentclass{article}
\usepackage[preprint]{neurips_2026}
\usepackage[utf8]{inputenc}
\usepackage[T1]{fontenc}
\usepackage{hyperref}
\usepackage{url}
\usepackage{booktabs}
\usepackage{amsmath,amsfonts}
\usepackage{graphicx}
\usepackage{longtable,array,calc}
\usepackage{etoolbox}

\providecommand{\pandocbounded}[1]{#1}
\newcolumntype{L}[1]{>{\raggedright\arraybackslash}p{#1}}
\AtBeginDocument{\setlength{\abovedisplayskip}{5pt}\setlength{\belowdisplayskip}{5pt}\setlength{\abovedisplayshortskip}{3pt}\setlength{\belowdisplayshortskip}{3pt}}
\title{Marginal Fidelity Does Not Establish User Simulation\\ in Demographic Synthetic Survey Panels:\\ Response Contracts, Support Collapse and Conditioning Failure}
\author{Alexander Doudkin\\Minds AI Labs, Inc.}

\begin{document}
\maketitle
\begin{abstract}
Demographic synthetic survey panels are often validated by matching
aggregate answers to a published survey. We test how little that
certificate establishes on six multiselect batteries from four survey
organisations in three countries. The headline analysis is restricted to
three instruments whose synthetic cohort and human target share the
stated population frame; one endogenous-base battery and two
frame-mismatched batteries remain visible as sensitivity analyses.

First, the response contract dominates measured fidelity. In the three
estimand-aligned instruments, committed sets leave 66 of 128
model-battery option slots empty in panels of up to 500 respondents,
versus 0 of 128 under per-option probability elicitation. Across the
eight uncapped model-instrument comparisons in that subset,
probabilities reduce option-marginal MAE by 4.53 to 7.30 points. The
capped instrument reverses on two models until the vectors are projected
onto its stated maximum. These are measurement effects: the human
targets are realised check-all responses, whereas the vectors are latent
inclusion propensities.

Second, published marginal agreement does not discriminate respondent
simulation from direct population estimation. On nine matched
model-battery pairs in the aligned subset, a no-persona
population-prevalence query averages 6.27 MAE versus 12.39 for committed
panels and is better on 9 of 9 pairs. Constraint-aware probability
vectors average 5.34 and are better than the query on 4 of 9, so the
baseline challenges the validation criterion rather than establishing
that direct estimation is uniformly best. On three unpublished
demographic cells, neither approach beats reciting the national
distribution. Exploratory thirty-cell analyses suggest that demographic
conditioning contains ordering signal with biased levels and that
offsets estimated from measured cells can reduce error, but the anchor
budgets and win rates are descriptive, post hoc, and not evidence of
out-of-sample superiority over survey estimators.

Population-marginal agreement is therefore a statement about an
elicitation contract and an estimand obtainable without simulated
respondents. It is not evidence of individual simulation.

\end{abstract}
\subsection{1. Introduction}\label{introduction}

A simulated survey respondent asked to check all that apply returns a
set. Asked instead, in a separate call, how likely each option is, the
same persona returns a probability vector. These are two \emph{response
contracts} over one persona specification. We study static synthetic
survey panels conditioned on four demographic attributes, not
interactive user simulators with histories, goals or memory; what may
generalize is the evaluation lesson, not the observed failure rate. The
paper's central result is that aggregate fidelity cannot by itself
validate individual simulation: a direct population query, with no
respondents, can outperform committed panels on the same
published-marginal endpoint (Section 6). This turns the
express-versus-simulate finding (Meister et al., 2025; Grief-Albert et
al., 2026) from a property of an estimator into a limitation of that
validation criterion. Elicitation changes what fidelity looks like, so
the benchmark measures a contract before it measures a simulator
(Sections 4 and 5); exploratory subgroup analyses then distinguish
demographic ordering signal from calibration of its levels (Section 7).

The first result is that the contract strongly determines how well these
demographic panels match a published survey. In the three instruments
with aligned population frames, whole options are returned by
\emph{nobody} in 66 of 128 \emph{model-battery option slots} under
committed sets, against zero under probabilities. On two of five models
no simulated United States respondent had visited an emergency room,
which 22.1 percent of real respondents report; on two of three, none of
the German cohort named immigration, at 23 percent. We call this
\textbf{support collapse}. It is an empirical, sample-size-dependent
diagnostic, not a proof of incapacity. Across the eight aligned uncapped
comparisons, probability elicitation improves marginal recovery by 4.53
to 7.30 points, while the capped battery shows why vectors must respect
the instrument's response constraint. \textbf{It is a better population
estimator on this endpoint, not necessarily a better respondent
simulator}: the human target is a realised check-all answer and the
vector is a latent propensity.

The second result is that agreement with published marginals does not
discriminate respondent simulation from direct population estimation. In
the aligned subset, a same-model population-prevalence query with no
simulated respondents, averaged over twenty repetitions, scores 6.27
against 12.39 for committed panels and is better on 9 of 9 matched
pairs. Constraint-aware probability vectors score 5.34 and beat the
query on 4 of 9, so the query is not a uniformly superior estimator. Its
role is diagnostic: because a non-simulation baseline can satisfy or
outperform a respondent panel on the usual marginal endpoint, that
endpoint cannot establish individual simulation. On held-out targets
with no published table, neither panel nor query beats reciting the
national table.

What the vectors recover of human joint structure beyond set size is
treated only exploratorily (Appendix A.8).

The contribution is not that probabilities can beat samples, which
Section 1.1's literature establishes. It is support collapse as a
multiselect failure diagnostic that needs no human data; a no-persona
baseline showing why marginal fidelity is not evidence of individual
simulation; and an exploratory decomposition of subgroup error into
ordering and level components, with a descriptive sparse-anchor
sensitivity analysis.

\subsubsection{1.1 Relation to prior work}\label{relation-to-prior-work}

Silicon sampling (Argyle et al., 2023; Aher et al., 2023) is often
evaluated by aggregate agreement with survey data, while audits document
collapsed within-group variance (Bisbee et al., 2024; Wang, Morgenstern,
and Dickerson, 2025), weak subgroup steerability (Santurkar et al.,
2023; von der Heyde, Haensch, and Wenz, 2025), instrument sensitivity
(Dominguez-Olmedo, Hardt, and Mendler-Dünner, 2024), and psychometric
invalidity (Lukauskas and Sarkauskaite, 2026). Interview-grounded agents
(Park et al., 2024) use much richer evidence than the four-attribute
panels studied here. Pilot anchoring and rectification are direct
precedents for Section 7 (Choi et al., 2026; Krsteski et al., 2026),
while specialised distribution estimators provide a stronger comparator
than zero-shot panels (Cao et al., 2025). A persistent Sim2Real gap
affects agentic evaluation outcomes (Seshadri et al., 2026; Zhou et al.,
2026), and verbalised confidence is protocol-sensitive (Kim and Kang,
2026). In the survey setting, concurrent work separates emulating
individuals from estimating populations (Grief-Albert et al., 2026), and
large method comparisons show that response generation changes alignment
(Ahnert et al., 2026); these studies address categorical or
single-select items. Multiselect differs because inclusion probabilities
need not sum to one and cardinality caps change what a probability
report means.

Three adjacent literatures bound our claims. Meister, Guestrin, and
Hashimoto (2025) show that LLMs describe categorical opinion
distributions better than they simulate them, and Zhang et al.~(2025)
that verbalizing a distribution mitigates the mode collapse of sampling
(Tian et al., 2023); ours is the multiselect case, where the estimand is
inclusion marginals plus set size and co-selection rather than one
simplex, and collapse takes the sharper form of options no respondent
expresses. SATA-Bench (Xu et al., 2025) documents count biases on
multiselect questions with known answers. In survey methodology,
forced-choice formats elicit more endorsements than check-all (Smyth et
al., 2006), and the probability readout is the graded limit of
forced-choice. First-token readouts do not extend to multiselect (Wang
et al., 2024) and stated probabilities diverge from realized choices
(Yamin et al., 2026), which is why the readout is a measurement
decision.

\subsection{2. Readouts, estimands, and what coverage
adds}\label{readouts-estimands-and-what-coverage-adds}

Let option inclusion be \(Y_{ik} \in \{0, 1\}\) for respondent \(i\) and
substantive option \(k\), with population marginal
\(p_k = \mathbb{E}\left[Y_{ik}\right]\) and set size
\(K_i = \sum_k Y_{ik}\). A set induces several non-equivalent population
estimands (inclusion marginals, cardinality, pairwise dependence,
constraint compliance, the distribution over sets), and better marginals
do not establish better sets. Our primary endpoint is option-marginal
MAE, \(\frac{1}{K}\sum_{k} \bigl|p_k^{S} - p_k^{H}\bigr|\), which says
nothing about which options co-occur or how many any one respondent
selects.

\subsubsection{2.1 Coverage as a prior
question}\label{coverage-as-a-prior-question}

MAE presumes the simulator produces the quantity being scored. Before
asking how far \(p_k^{S}\) is from \(p_k^{H}\), ask whether \(p_k^{S}\)
is nonzero at all. Define an option \textbf{zero-coverage} under a
readout at sample size \(n\) when no respondent expresses it, and
\textbf{degenerate} when every respondent returns the same value, so
between-respondent variance is zero. Both are threshold-free and
computable from simulator output alone. Their cost in human terms is the
summed human prevalence of the zero-coverage options,
\textbf{unexpressed human prevalence}, a sum of option prevalences
rather than a population share. Coverage is prior to calibration and not
implied by it: a panel can post a respectable average error while
emitting nothing on a third of its options, because a collapsed
low-prevalence option costs little absolute error.

Two properties qualify the statistic. It is empirical and
\(n\)-dependent: zero selections in 500 respondents is not
impossibility. And the comparison is asymmetric by construction: a
continuous readout avoids zero coverage merely by staying positive, and
the smallest option mean ours returns on a national battery is 1.70
percent. Zero holes under the probability readout therefore evidence
\textbf{marginal expressivity}, not joint behavioural support: a persona
reporting 70/20/5/5 has an expressive vector while its realized
behaviour may still be \(\{A\}\) every time.

We compare two readouts head to head, direct multiselect (one realized
set) and raw option probabilities (one 0-100 inclusion forecast per
substantive option, averaged after division by 100 with no projection or
calibration). Appendix E gives the psychological motivation; it is not
an identified mechanism.

\subsection{3. Design}\label{design}

Every study below runs on named language models through provider APIs.
The archived records contain model-output text or parsed terminal
responses, validity flags, identifiers and requested decoding settings
where the runner retained them; they are not complete raw API envelopes,
and token usage is run-level or absent for some collections. No
commercial simulation platform contributes to the reported model
outputs.

\textbf{Cohorts.} A cohort is 500 demographic-only personas. The
retained allocation audits show that the larger demographic files match
specified \emph{one-way marginals} to 0.1 points; because no source-safe
relationships were supplied, the generator used seeded independent
assignment rather than official four-way joint cells. The runs use the
first 500 rows of each file, a convenience slice whose one-way shares
differ from the audit targets by up to 5.4 percentage points (the two US
files 5.4 and 3.2, UK/GB 4.2, DE 5.0;
\texttt{analysis/cohort\_deviation.py}). A sensitivity cohort
reallocated 500 rows over the empirical profile cells of the generated
US file and moves one paired contract difference from \(+7.26\) to
\(+7.43\) with unchanged coverage (Appendix H.3); it does not validate
the source population's joint distribution. At answer time a persona is
a system prompt naming four attributes and nothing else. United States
marginal targets are derived from CES 2024 (gender, age band, Census
region, education); UK/GB files use gender, age band, nation and social
grade; the German file uses gender, age band, region and ISCED
education.

\textbf{Batteries and target scope.} Six multiselect batteries come from
four survey organisations in three countries: CES 2024 CC24\_300 (media
use, 5 options), CC24\_300d (political activity on social media, 6
options, social-media-user base), and CC24\_305 (life events, 13
options); Ofcom Adult Media Literacy item IN5A (online activities, 9
options); the ONS July 2025 cost-of-living item (11 options); and
Standard Eurobarometer 103 QA3 for Germany (16 options, maximum two
selections). The headline subset is CC24\_300, CC24\_305 and QA3.
CC24\_300d is sensitivity-only because eligibility is defined from the
model's own committed response and 131 valid gpt-4o-mini arrays are
empty; IN5A is sensitivity-only because its human base is internet users
but the synthetic cohort represents all UK adults; ONS is
sensitivity-only because its target is Great Britain but the synthetic
roster includes 14 Northern Ireland personas. The release includes
target-file SHA-256 digests, which authenticate the shipped bytes but do
not establish when they were created. Appendix H gives sources, bases,
the translation caveat and these exclusions.

\textbf{Readouts.} Each battery is asked twice of each persona, in
independent calls. The \textbf{direct} readout asks for a JSON array of
the options that apply, at temperature 1.0, validated against the
released option set. The \textbf{probability} readout asks for one 0-100
inclusion forecast per option, at temperature 0.2, aggregated by
dividing by 100 and averaging with no projection, normalization,
calibration or repair. Invalid output is re-asked at most twice; the
capped battery additionally rejects answers exceeding the cap. Each
contract runs at its conventional decoding temperature, 1.0 for a
sampled commitment and 0.2 for a numeric estimate, so we also ran the
committed-set contract at 0.2. Matching temperature moves the contract
difference by +0.18 points on average, or \(-0.08\) with the other of
two collections of the CC24\_305 cell, and changes no sign (Appendix
B.2). The persona system prompt reads: ``You are simulating one specific
{[}population{]} survey respondent. Characteristics: {[}four
attributes{]}. Answer exactly as this specific person would, based on
what is typical and plausible for someone with these characteristics.
Answer only in the requested format.'' Appendix B.6 ablates the
typicality clause.

\textbf{Scoring.} The endpoint for readout \(c\) on battery \(b\) is
\(\mathrm{MAE}_c(b) = |\mathcal{O}_b|^{-1}\sum_o |\hat{q}^{c}_{o} - q_o|\)
against the released shares \(q_o\), and the coverage statistics of
Section 2.1 are computed on the same respondent-by-option matrices. A
persona is four attributes, so 500 respondents realise only 57 to 120
distinct prompts and repeated draws from one prompt are close to
deterministic; the unit of resampling for the national-battery
comparison is therefore the distinct prompt. Those intervals are paired
bootstraps that resample distinct prompts with replacement, carrying
each prompt's respondents with it, and recompute both readouts on the
same draw (2,000 replicates, seed fixed and reported). The respondent
bootstrap, a median 2.27 times narrower, is in Appendix C. The
fixed-cell held-out intervals of Section 6 instead resample repeated
calls within one prompt and quantify decoding noise alone. Intervals are
conditional on one cohort, prompt set and collection window; the model
axis is addressed by running every battery on several models.

\subsection{4. Support collapse under committed-set
elicitation}\label{support-collapse-under-committed-set-elicitation}

Coverage is computed from the response artifacts alone
(\texttt{analysis/headline\_coverage.py}), using the human targets only
for the prevalence figures below. \textbf{Every national run has zero
zero-coverage options under the probability readout}, so only the
committed-set side varies, from one option of five to ten of sixteen;
Appendix G gives the per-run table.

In the estimand-aligned headline subset, \textbf{66 of 128 option slots
draw zero selections in panels of up to 500 respondents under the
committed-set readout, and 0 of 128 under the probability readout},
which a continuous readout achieves almost by construction. The 66 are
descriptive observed zeros, not hypothesis-test rejections. Across all
six instruments, including the three sensitivity-only batteries, the
corresponding counts are 85 of 206 and 0 of 206; Appendix G reports
every run and a heuristic sampling-surprise screen that is not used
inferentially. Counting saturated options as well, the all-instrument
committed-set total is 100 of 206 slots with zero between-respondent
variance, against 0 of 206 under probabilities.

The unexpressed options are not obscure. Two of five models returned
nobody who had visited an emergency room (human 22.1 percent), and on
the German battery none of the three models' respondents named
government debt (12.0) and two of three none named immigration (23.0).
On the life-events battery no model's respondents included anyone who
had married, had a child or been a victim of a crime (3.7, 2.5 and 3.9
percent). A single German run leaves 108.0 points of summed human
prevalence unexpressed. These observations motivate the diagnostic;
dependence among repeated demographic prompts prevents treating them as
independent binomial trials.

Two qualifications from Section 2.1 apply: the statistic is empirical
and \(n\)-dependent, and the comparison is asymmetric by construction,
so the probability column evidences marginal expressivity rather than
joint behavioural support. It is not induced by the typicality clause of
the persona instruction: without it, or asking for one randomly drawn
person, the zero-coverage count is unchanged, though the readout effect
on that battery shrinks by a third (Appendix B.6). Coverage is also
necessary, not sufficient: Section 5 reports one battery where the
probability readout is substantially \emph{worse} on error despite
perfect coverage.

\subsection{5. The readout effect and its
boundary}\label{the-readout-effect-and-its-boundary}

Twenty battery-model runs completed. Across all instruments, 19,510
records were collected of 20,000 planned and 19,410 passed the archived
validators; the 490 structurally absent records are CC24\_300d personas
outside its model-defined social-media base. The headline subset
contains eleven model-battery runs: three on CC24\_300, five on
CC24\_305 and three on QA3. Differences are \(\Delta\) with paired
bootstrap intervals clustered by distinct prompt (2,000 replicates, seed
20260901; Appendix C). These intervals are conditional on the retained
cohort and prompts, not uncertainty over populations, models or
instruments.

{\def\LTcaptype{none} 
{\footnotesize\begin{longtable}[]{@{}
  >{\raggedright\arraybackslash}p{(\linewidth - 12\tabcolsep) * \real{0.1429}}
  >{\raggedright\arraybackslash}p{(\linewidth - 12\tabcolsep) * \real{0.1429}}
  >{\raggedright\arraybackslash}p{(\linewidth - 12\tabcolsep) * \real{0.1429}}
  >{\raggedright\arraybackslash}p{(\linewidth - 12\tabcolsep) * \real{0.1429}}
  >{\raggedright\arraybackslash}p{(\linewidth - 12\tabcolsep) * \real{0.1429}}
  >{\raggedright\arraybackslash}p{(\linewidth - 12\tabcolsep) * \real{0.1429}}
  >{\raggedright\arraybackslash}p{(\linewidth - 12\tabcolsep) * \real{0.1429}}@{}}
\toprule\noalign{}
\begin{minipage}[b]{\linewidth}\raggedright
Battery
\end{minipage} & \begin{minipage}[b]{\linewidth}\raggedright
Status
\end{minipage} & \begin{minipage}[b]{\linewidth}\raggedright
Opts
\end{minipage} & \begin{minipage}[b]{\linewidth}\raggedright
Models
\end{minipage} & \begin{minipage}[b]{\linewidth}\raggedright
Direct MAE
\end{minipage} & \begin{minipage}[b]{\linewidth}\raggedright
Probability MAE
\end{minipage} & \begin{minipage}[b]{\linewidth}\raggedright
Difference
\end{minipage} \\
\midrule\noalign{}
\endhead
\bottomrule\noalign{}
\endlastfoot
CES CC24\_300 media use (US) & headline & 5 & 3 & 8.72--16.80 &
3.97--9.55 & +4.53 to +7.26 \\
CES CC24\_305 life events (US) & headline & 13 & 5 & 8.84--11.81 &
2.15--5.31 & +5.18 to +7.30 \\
EB103 QA3 issues, max-2 cap (DE) & headline & 16 & 3 & 11.38--16.25 &
6.68--35.48 & -20.46 to +4.70 \\
CES CC24\_300d political SM (US)\textsuperscript{a} & sensitivity & 6 &
3 & 16.30--24.91 & 4.52--11.59 & +4.70 to +20.39 \\
Ofcom IN5A online activities\textsuperscript{b} & sensitivity & 9 & 3 &
16.76--20.89 & 4.38--6.90 & +12.38 to +14.01 \\
ONS cost-of-living\textsuperscript{c} & sensitivity & 11 & 3 &
23.07--33.96 & 8.74--23.53 & +10.43 to +14.33 \\
\end{longtable}}
}

\textsuperscript{a} Eligibility is the persona's own CC24\_300 committed
response and differs by model (449, 324, 482). On gpt-4o-mini, 131 of
324 follow-up arrays are empty. Treating those as literal empty sets
gives the table's +4.70; mapping them to the explicit ``None of the
above'' option gives +1.57. Neither coding is headline evidence
(Appendix H.1). \textsuperscript{b} The target base is UK internet
users; the synthetic cohort is all UK adults. \textsuperscript{c} The
target is Great Britain; the synthetic cohort includes 14 Northern
Ireland personas. Absolute comparisons for \textsuperscript{b} and
\textsuperscript{c} are not population-matched.

\textbf{Depth on one battery.} CC24\_305 ran on five models across three
providers: all reproduce the direction, \(+5.18\) to \(+7.30\), every
clustered interval excluding zero (Appendix B.1), and the effect
survives a matched readout-by-temperature two-by-two (Appendix B.2).

\textbf{The boundary, and it is a reversal rather than a null.} All
eight uncapped comparisons in the aligned subset favour probabilities,
from \(+4.53\) to \(+7.30\) points. The capped battery breaks this, and
it is also the one instrument with unofficially translated option
labels, so constraint and translation are confounded. On Eurobarometer
QA3, claude-haiku-4-5 gives \(-10.50\) \([-11.38, -9.12]\) and
gpt-4o-mini \(-20.46\) \([-21.33, -19.03]\), while gemini-3.7-flash
gives \(+4.70\). The elicited vectors sum to 620 and 760 points on the
two reversing models against a human 192; scaling each vector to an
admissible total changes the first two differences to \(+12.76\) and
\(+9.79\) (Appendix A.5). This post hoc projection shows that
unconstrained mass explains the observed reversal; it does not isolate a
causal cap effect.

Two descriptive checks bound alternative explanations without turning
the model runs into population-level inference. First, the committed-set
readout has a binary-sampling floor; at the retained distinct-prompt
counts and target prevalences it is 1.9 to 4.3 points per battery, while
the smallest aligned uncapped effect is 4.53 on a battery whose
estimated floor is 2.9 (Appendix D). Second, a heuristic calculation of
\((1-p_H)^{n_{\mathrm{prompt}}}\) puts 18 of the 85 all-instrument zeros
at or above 0.05 and 34 below \(10^{-3}\) (Appendix G). Repeated prompts
are dependent and the human prevalence is treated as fixed, so these
values are diagnostics rather than calibrated tests and no FDR result
supports the headline. Recall probes also fail to establish absence of
contamination: models decline literal table requests, but
retrieved-and-adjusted priors remain possible (Appendix B.4).

The defensible summary is therefore a same-direction effect on two
aligned uncapped instruments, replicated across eight model-instrument
runs, plus a constraint-sensitive boundary on one capped instrument. We
do not meta-analyse the batteries: model spread is not a sampling
variance and the instruments are not a random sample from a defined
population. An all-instrument option-level regression is retained as
exploratory description only (Appendix A.2).

\textbf{Set size, the one joint quantity the marginals imply.} The
target marginals fix the human mean set size, and each contract implies
one: the mean committed-set size and the summed mean probabilities under
an independent reading of the vector. In the aligned uncapped subset,
probabilities are closer in 6 of 8 runs and committed sets in 2; on the
capped battery the raw vectors imply 1.2 to 3.9 times the human set size
(Appendix A.7). Marginal MAE and respondent-level structure are
different estimands.

\subsection{6. What marginal agreement does not
establish}\label{what-marginal-agreement-does-not-establish}

Two tests ask what agreement with a published marginal certifies.

\textbf{A held-out contamination control.} Every national target is
public, so recall cannot be excluded by design. We therefore computed
unpublished weighted CC24\_305 marginals for three four-way demographic
cells. The retained study record says the target file and selection rule
were fixed before these calls; the shipped digest authenticates the file
now but cannot independently prove that chronology. Reciting the
national table instead of conditioning on the cell costs 7.00 points on
average. Two models answered as 500 repeated draws per fixed cell under
both contracts. Because each cell is one prompt specification, its
intervals quantify decoding variation, not independent respondents or
population uncertainty.

The readout effect survives: five of six cell-model comparisons favour
probabilities, from \(+1.88\) to \(+12.05\), every interval excluding
zero, and direct-set zero coverage is worse than on the national targets
(6 to 11 of 13 options). Paraphrasing every option label leaves five of
six favouring probabilities, effect sizes correlating at \(r = 0.94\)
with the originals. Two results cut the other way. On one cell
gpt-4o-mini reverses, with the direct readout better by \(2.77\)
\([-3.26, -2.29]\). Appendix I traces it to the paper's own rare-option
mechanism; under paraphrase it shrinks to \(-0.25\). And the probability
readout is not immune to collapse: on one cell-model pair it returns
exactly zero on 8 of 13 options. The main result is negative and larger
than anything the readout addresses. On these unpublished targets
\emph{neither} readout beats the trivial baseline of reciting the
national table (the better of the two contracts, averaged over the six
cell-model pairs, 7.65; no-persona query 7.47; national table 7.00;
three of six pairs each way). Re-estimating each cell's targets on a
random half of its respondents, which removes the winner's curse in the
cell selection, does not rescue the simulator: the national table then
scores 6.97 against the panel's best readout at 7.52, splitting the six
cell-model pairs three and three. On published national batteries this
failure is invisible, because the national answer is the right answer,
and validation against published marginals alone cannot detect it.
Subgroup conditioning, not readout choice, is the binding constraint off
the national table; Section 7 takes it apart (Appendix I).

\textbf{The no-persona baseline, and where it stops working.} One prompt
asks the same model for each option's population prevalence, with no
personas, and averages up to twenty valid repetitions. Across nine
matched pairs in the aligned subset, the query scores 6.27 against 12.39
for committed panels and is better on 9 of 9. A single repetition
averages 6.45 and is also better on 9 of 9. Constraint-aware probability
vectors average 5.34 and beat the query on 4 of 9, so the baseline does
not establish that direct estimation is best. It establishes the
narrower point: published-marginal agreement is obtainable without
simulated respondents and therefore cannot certify individual
simulation. On held-out cells the query wins three of six comparisons
against the better panel readout and both average worse than reciting
the national table. The query produces no respondent sets, cardinality
or co-selection. Appendix B.3 retains the all-instrument 18-pair
sensitivity summary, labelled with the CC24\_300d and population-frame
limitations. We make no training-cutoff claim from the gpt-4o-mini
identifier because its exact dated resolution was recorded after
collection (Appendix B.4).

\subsection{7. Exploratory decomposition of subgroup
calibration}\label{exploratory-decomposition-of-subgroup-calibration}

Section 6 leaves the panel behind a lookup table on three unpublished
cells. We use thirty further CES cells on two models to describe where
that error lies and to explore a simple offset correction (Appendix J).
These analyses were specified after the initial three-cell result and
are not confirmatory comparisons with survey estimators.

Against cell truth the probability contract scores 6.08 and 5.09 and the
committed-set contract 11.36 and 12.14, all worse than the 4.85 from
reciting the national table. Writing each cell's option share as a
deviation from that option's mean, simulated and human deviations have
per-option median correlations of \(r = 0.79\) and \(0.86\) (pooled 0.54
and 0.83). Because cells were selected to span the human divergence
range, those correlations are descriptive upper bounds. Mean absolute
per-option level bias is 4.62 and 4.48 points. Under committed sets only
4 and 5 of 13 options vary across cells at all.

That diagnosis predicts a repair: one offset per option, estimated on
\(k\) known cells and applied to held-out cells, compared against the
national table anchored on the same cells by the same procedure.

{\def\LTcaptype{none} 
{\footnotesize\begin{longtable}[]{@{}
  >{\raggedright\arraybackslash}p{(\linewidth - 8\tabcolsep) * \real{0.2000}}
  >{\raggedright\arraybackslash}p{(\linewidth - 8\tabcolsep) * \real{0.2000}}
  >{\raggedright\arraybackslash}p{(\linewidth - 8\tabcolsep) * \real{0.2000}}
  >{\raggedright\arraybackslash}p{(\linewidth - 8\tabcolsep) * \real{0.2000}}
  >{\raggedright\arraybackslash}p{(\linewidth - 8\tabcolsep) * \real{0.2000}}@{}}
\toprule\noalign{}
\begin{minipage}[b]{\linewidth}\raggedright
Anchor cells
\end{minipage} & \begin{minipage}[b]{\linewidth}\raggedright
Probability, anchored
\end{minipage} & \begin{minipage}[b]{\linewidth}\raggedright
Committed set, anchored
\end{minipage} & \begin{minipage}[b]{\linewidth}\raggedright
National table
\end{minipage} & \begin{minipage}[b]{\linewidth}\raggedright
National, anchored alike
\end{minipage} \\
\midrule\noalign{}
\endhead
\bottomrule\noalign{}
\endlastfoot
none & 6.08 / 5.09 & 11.36 / 12.14 & 4.85 & --- \\
1 & 5.62 / 3.52 & 8.32 / 7.91 & 4.85 & 6.49 \\
5 & 4.55 / 2.90 & 7.53 / 7.12 & 4.84 & 5.06 \\
10 & 4.42 / 2.77 & 7.45 / 6.99 & 4.87 & 4.98 \\
\end{longtable}}
}

Entries are haiku-4-5 / gemini-3.7-flash; anchors drawn at random,
scored only on held-out cells, 200 draws per budget; the anchor draws
are shared across models, so the national columns are identical for
both.

In these same data, ten anchors reduce haiku's probability MAE from 6.08
to 4.42 and one reduces gemini's from 5.09 to 3.52; the corresponding
empirical draw-win rates against the unadjusted national table are 95
and 96 percent. These are resampling summaries conditional on selected
cells, models and a post hoc correction, not success probabilities for
new instruments or populations. The committed-set correction remains
above the national table at every displayed budget. Applying the same
offsets to a cell-invariant national table adds noise, so the
anchored-table comparison is informative but not a substitute for a
fitted survey-statistical baseline.

The same descriptive pattern appears on a second CES battery across four
models: probability outputs preserve cell ordering more than committed
sets, and offsets reduce their MAE. Anchor counts reaching a 95-percent
empirical draw-win rate range from two to ten, but this is not a
preregistered budget guarantee. Two models already beat the national
table before anchoring. A written profile generated from the same four
attributes also reduces level bias on one model and battery, without
adding person-specific evidence. All three analyses reuse CES microdata,
compare against a deliberately weak table baseline, and omit the
small-area, regression and multilevel estimators a survey methodologist
would normally fit; their role is hypothesis generation (Appendix
J.2--J.3).

\subsection{8. Conclusion}\label{conclusion}

For static demographic synthetic survey panels, measured fidelity
changes sharply with the response contract, and matching population
marginals cannot establish individual simulation. In the aligned subset,
committed sets leave 66 of 128 model-option slots empty while
probabilities leave none; probabilities reduce marginal MAE on all eight
uncapped comparisons, but must be projected to respect a capped
instrument. A no-persona query beats committed panels on all nine
aligned matched pairs, though not the constraint-aware probability
vectors on average. Exploratory subgroup analyses suggest ordering
signal with biased levels, but do not establish an anchor budget or
superiority to survey estimators. Evaluate a synthetic user on the
individual and joint estimands it is claimed to simulate; marginal
agreement is evidence about a readout, not a person.

\subsection{9. Limitations and ethics}\label{limitations-and-ethics}

\textbf{Targets differ in mode and date.} The six instruments were
fielded by four organisations in different years and by different modes,
and mode and format move check-all endorsement by several points (Smyth
et al., 2006), the same order as the effects measured here. That does
not threaten the contract comparison, which is within-instrument and
paired, but absolute MAE levels are not comparable across batteries and
no single number here is a fidelity score for a survey population.

\textbf{The intervals are conditional.} The twenty national-battery
intervals resample the distinct prompt, the effective unit at 57 to 120
prompts per cohort, and are a median 2.27 times wider than the
respondent bootstrap the literature reports; all twenty exclude zero
under either (Appendix C). They are conditional prompt-clustered
bootstraps in the sense of Section 3, not inferential statements across
models, prompts or time. The fixed-cell held-out intervals instead
resample repeated calls within one prompt and quantify decoding noise
alone (Appendix I). All batteries were collected in one window, so time
effects are not separable from model effects. Absolute levels are less
stable than the paired contrasts: re-collecting CC24\_305 on
claude-haiku-4-5 moves absolute MAE by about a point under both
contracts and the paired readout effect by 0.064 (Appendix B.2), and the
probability readout is the more stable per option. Whether that holds on
the other five instruments is untested. Four of the twenty runs have
fewer than 500 valid responses under one contract; point estimates use
every valid record and intervals pair by respondent identifier on the
records valid under both (Appendix B.5). The committed-set binomial
floor (Appendix D) is at most 1.8 points at \(n = 500\) and 3.7 to 5.3
at the effective \(n\) of 120 to 57 distinct prompts, both at 50 percent
prevalence; at the sealed prevalences the per-battery floors are 1.9 to
4.3, and the smallest uncapped effect, 4.53 on CC24\_300, exceeds that
battery's 2.9. The floor does not affect the coverage statistic.

\textbf{Grounding is varied once, narratively.} Every headline persona
is four demographic attributes. A post hoc written-profile arm generated
from those same attributes lowers error on one model and battery, but
adds no observed information about an individual and supplies no
confirmatory evidence about grounding. What interviews, histories, goals
or persistent interaction would recover is untested.

\textbf{Three limitations deserve the most weight.} First, the two
readouts are separate calls, so this is a comparison between two
elicitations of one demographic prompt; within-call consistency lies
outside the design. Second, the endpoint and treatment do not measure
the same behaviour. Human targets are realised check-all answers, while
probability elicitation asks for latent per-option propensities. Better
marginal recovery can therefore accompany less faithful respondent-level
behaviour. Third, target contamination cannot be excluded. The release
digests authenticate the shipped targets but do not prove collection
chronology, analyst blinding or model ignorance. Recall probes do not
establish absence of retrieved priors, all six items or earlier waves
are public, and exact dated-model provenance was not captured
contemporaneously for every call. The unpublished-cell and paraphrase
analyses reduce literal-table explanations but do not provide a
genuinely private, novel instrument.

\textbf{Scope and population frames.} Headline claims concern CC24\_300,
CC24\_305 and QA3 only. CC24\_300d has endogenous eligibility and
ambiguous empty arrays; IN5A compares an all-adult synthetic cohort with
an internet-user target; ONS compares a roster containing Northern
Ireland with a Great Britain target. Their rows are retained as
sensitivity evidence, not pooled or used for absolute
panel-versus-target claims. The readout effect is also prompt-sensitive:
two CC24\_305 prompt variants shrink the paired difference from
\(+4.76\) to \(+3.31\) and \(+3.15\) on one model. The German labels are
unofficial translations. The anchor analysis lacks a survey-statistical
comparator and is post hoc. Set size is implied from marginals, realised
co-selection is exploratory on CES only, and exact-set recovery remains
outside the design.

The authors have a commercial interest in synthetic-panel simulation. We
disclose it because the paper's central results argue against the way
such systems are usually validated, and because that direction of
interest is the relevant one for a reader weighing them. No product of
ours is evaluated here: every system measured is a third-party language
model reached through its provider's public API. No new human
participants were recruited; all human data are published aggregates or
public microdata used under their terms. Synthetic profiles must not be
framed as people or substitutes for lived experience (Agnew et al.,
2024). The release package contains target files with current digests,
cohort-allocation audits, structured model-response records, and
analysis scripts; it contains no human free text, names, or private
identifiers.

\section*{References}
\begingroup\footnotesize\setlength{\parindent}{0pt}\setlength{\parskip}{3pt}
Agnew, William, Bergman, A. Stevie, Chien, Jennifer, D\'iaz, Mark, El-Sayed, Seliem, Pittman, Jaylen, Mohamed, Shakir, McKee, Kevin R. (2024). The Illusion of Artificial Inclusion. Proceedings of the 2024 CHI Conference on Human Factors in Computing Systems.

Aher, Gati V., Arriaga, Rosa I., Kalai, Adam Tauman (2023). Using Large Language Models to Simulate Multiple Humans and Replicate Human Subject Studies. Proceedings of the 40th International Conference on Machine Learning 202, 337--371.

Ahnert, Georg, Haensch, Anna-Carolina, Plank, Barbara, Strohmaier, Markus (2026). Survey Response Generation: Generating Closed-Ended Survey Responses In-Silico with Large Language Models. Proceedings of the 64th Annual Meeting of the Association for Computational Linguistics (Volume 1: Long Papers), 41554--41577.

Argyle, Lisa P., Busby, Ethan C., Fulda, Nancy, Gubler, Joshua R., Rytting, Christopher, Wingate, David (2023). Out of One, Many: Using Language Models to Simulate Human Samples. Political Analysis 31(3), 337--351.

Bisbee, James, Clinton, Joshua D., Dorff, Cassy, Kenkel, Brenton, Larson, Jennifer M. (2024). Synthetic Replacements for Human Survey Data? The Perils of Large Language Models. Political Analysis 32(4), 401--416.

Cao, Yong, Liu, Haijiang, Arora, Arnav, Augenstein, Isabelle, Röttger, Paul, Hershcovich, Daniel (2025). Specializing Large Language Models to Simulate Survey Response Distributions for Global Populations. Proceedings of the 2025 Conference of the Nations of the Americas Chapter of the Association for Computational Linguistics: Human Language Technologies (Volume 1: Long Papers), 3141--3154.

Choi, Eun Cheol, Kim, Youngrae, Pugalenthi, Prabhu, Chen, Hong-En, Huang, Bo-Ruei (2026). Beyond the Mean: Three-Axis Fidelity for Aligning LLM-Based Survey Simulators from Small Pilot Data. arXiv preprint arXiv:2606.28963.

Dominguez-Olmedo, Ricardo, Hardt, Moritz, Mendler-Dunner, Celestine (2024). Questioning the Survey Responses of Large Language Models. arXiv preprint arXiv:2306.07951.

European Commission (2025). Standard Eurobarometer 103 (EB 103.3), Spring 2025. European Commission.

Grief-Albert, Seth, Bo, Jessica, Jiao, Difan, Anderson, Ashton (2026). Emulate or Estimate? The Divergent Strengths of Base and Post-Trained Language Models for Opinion Simulation. arXiv preprint arXiv:2608.03044.

Kim, Hankyeol, Kang, Pilsung (2026). Same Answer, Different Confidence: Protocol Sensitivity in LLM Confidence Calibration. arXiv preprint arXiv:2605.27752.

Krosnick, Jon A. (1991). Response Strategies for Coping with the Cognitive Demands of Attitude Measures in Surveys. Applied Cognitive Psychology 5(3), 213--236.

Krsteski, Stefan, Russo, Giuseppe, Chang, Serina, West, Robert, Gligori\'c, Kristina (2026). Valid Survey Simulations with Limited Human Data: The Roles of Prompting, Fine-Tuning, and Rectification. Proceedings of the 64th Annual Meeting of the Association for Computational Linguistics (Volume 1: Long Papers), 10887--10906.

Lichtenstein, Sarah, Slovic, Paul (1971). Reversals of Preference Between Bids and Choices in Gambling Decisions. Journal of Experimental Psychology 89(1), 46--55.

Lukauskas, Mantas, \v Sarkauskait\.e, Viktorija (2026). Plausible but Not Valid: A Psychometric Audit of LLMs as Synthetic Survey Respondents. arXiv preprint arXiv:2608.14606.

Manski, Charles F. (2004). Measuring Expectations. Econometrica 72(5), 1329--1376.

Meister, Nicole, Guestrin, Carlos, Hashimoto, Tatsunori (2025). Benchmarking Distributional Alignment of Large Language Models. Proceedings of the 2025 Conference of the Nations of the Americas Chapter of the Association for Computational Linguistics: Human Language Technologies (Volume 1: Long Papers), 24--49.

Ofcom (2025). Adults' Media Use and Attitudes 2025: Adult Media Literacy Core Survey 2024 data tables. Ofcom.

Office for National Statistics (2025). Public opinions and social trends, Great Britain: household finances, 2 to 27 July 2025. Office for National Statistics.

Park, Joon Sung, Zou, Carolyn Q., Kamphorst, Jonne, Egan, Niles, Shaw, Aaron, Hill, Benjamin Mako, Cai, Carrie, Morris, Meredith Ringel, Liang, Percy, Willer, Robb, Bernstein, Michael S. (2024). LLM Agents Grounded in Self-Reports Enable General-Purpose Simulation of Individuals. arXiv preprint arXiv:2411.10109.

Santurkar, Shibani, Durmus, Esin, Ladhak, Faisal, Lee, Cinoo, Liang, Percy, Hashimoto, Tatsunori (2023). Whose Opinions Do Language Models Reflect?. Proceedings of the 40th International Conference on Machine Learning 202, 29971--30004.

Schaffner, Brian F., Ansolabehere, Stephen, Shih, Marissa (2025). Cooperative Election Study Common Content, 2024. Harvard Dataverse.

Schwarz, Norbert (1999). Self-Reports: How the Questions Shape the Answers. American Psychologist 54(2), 93--105.

Seshadri, Preethi, Cahyawijaya, Samuel, Odumakinde, Ayomide, Singh, Sameer, Goldfarb-Tarrant, Seraphina (2026). Lost in Simulation: LLM-Simulated Users are Unreliable Proxies for Human Users in Agentic Evaluations. arXiv preprint arXiv:2601.17087.

Slovic, Paul (1995). The Construction of Preference. American Psychologist 50(5), 364--371.

Smyth, Jolene D., Dillman, Don A., Christian, Leah Melani, Stern, Michael J. (2006). Comparing Check-All and Forced-Choice Question Formats in Web Surveys. Public Opinion Quarterly 70(1), 66--77.

Tian, Katherine, Mitchell, Eric, Zhou, Allan, Sharma, Archit, Rafailov, Rafael, Yao, Huaxiu, Finn, Chelsea, Manning, Christopher D. (2023). Just Ask for Calibration: Strategies for Eliciting Calibrated Confidence Scores from Language Models Fine-Tuned with Human Feedback. Proceedings of the 2023 Conference on Empirical Methods in Natural Language Processing, 5433--5442.

Tourangeau, Roger, Rips, Lance J., Rasinski, Kenneth A. (2000). The Psychology of Survey Response. Cambridge University Press.

Tversky, Amos, Sattath, Shmuel, Slovic, Paul (1988). Contingent Weighting in Judgment and Choice. Psychological Review 95(3), 371--384.

von der Heyde, Leah, Haensch, Anna-Carolina, Wenz, Alexander (2025). Vox Populi, Vox AI? Using Large Language Models to Estimate German Vote Choice. Social Science Computer Review 44(3), 549--571.

Wang, Xinpeng, Ma, Bolei, Hu, Chengzhi, Weber-Genzel, Leon, Röttger, Paul, Kreuter, Frauke, Hovy, Dirk, Plank, Barbara (2024). "My Answer is C": First-Token Probabilities Do Not Match Text Answers in Instruction-Tuned Language Models. Findings of the Association for Computational Linguistics: ACL 2024, 7407--7416.

Wang, Angelina, Morgenstern, Jamie, Dickerson, John P. (2025). Large Language Models That Replace Human Participants Can Harmfully Misportray and Flatten Identity Groups. Nature Machine Intelligence 7, 400--411.

Xu, Weijie, Cui, Shixian, Fang, Xi, Xue, Chi, Eckman, Stephanie, Reddy, Chandan K. (2025). SATA-BENCH: Select All That Apply Benchmark for Multiple Choice Questions. arXiv preprint arXiv:2506.00643.

Yamin, Khurram, Tang, Jingjing, Cortes-Gomez, Santiago, Sharma, Amit, Horvitz, Eric, Wilder, Bryan (2026). When Agents Say One Thing and Do Another: Validating Elicited Beliefs from LLMs. arXiv preprint arXiv:2602.06286.

Zhang, Jiayi, Yu, Simon, Chong, Derek, Sicilia, Anthony, Tomz, Michael R., Manning, Christopher D., Shi, Weiyan (2025). Verbalized Sampling: How to Mitigate Mode Collapse and Unlock LLM Diversity. arXiv preprint arXiv:2510.01171.

Zhou, Xuhui, Sun, Weiwei, Ma, Qianou, Xie, Yiqing, Liu, Jiarui, Du, Weihua, Welleck, Sean, Yang, Yiming, Neubig, Graham, Wu, Sherry Tongshuang, Sap, Maarten (2026). Mind the Sim2Real Gap in User Simulation for Agentic Tasks. arXiv preprint arXiv:2603.11245.

\endgroup

\newpage
\subsection{Appendix A. Option-level and sensitivity
analyses}\label{appendix-a.-option-level-and-sensitivity-analyses}

The retained study record identifies option-marginal MAE per contract
with paired prompt-cluster bootstrap intervals as the battery-level
endpoint. The shipped hashes authenticate current files but cannot prove
that this specification preceded collection. Analyses in this appendix
are explanatory or sensitivity analyses and do not establish
population-level effects.

\subsubsection{A.1 Option-level dataset}\label{a.1-option-level-dataset}

The option-level file has one row per option per battery per model: 206
rows over 60 distinct survey options, since each battery is evaluated on
three to five models. Analyses that treat an option as the unit and pool
across models, including the regression of A.2, use all 206 rows and
cluster nothing, which is stated where it matters. Counts of
\emph{distinct options} are 60; counts of \emph{evaluated slots} are
206, and the paper uses the latter phrase wherever the number appears.

\subsubsection{A.2 Where the advantage comes
from}\label{a.2-where-the-advantage-comes-from}

Let \(e_{o,c}\) be the absolute option-marginal error of option \(o\)
under readout \(c\), \(d_o = e_{o,\mathrm{dir}} - e_{o,\mathrm{prob}}\)
the per-option improvement, and \(\tilde{\ell}_o\) centred log human
prevalence. Over all 206 option rows, a descriptive regression of
improvement on centred log prevalence has slope \(\hat{\beta}=7.13\). We
omit an inferential standard error and test because rows repeat options
across models, include sensitivity-only population frames, and are not
sampled independently. Descriptive improvement by prevalence tercile is:

{\def\LTcaptype{none} 
{\footnotesize\begin{longtable}[]{@{}ll@{}}
\toprule\noalign{}
Tercile & Mean improvement (pp) \\
\midrule\noalign{}
\endhead
\bottomrule\noalign{}
\endlastfoot
Rare & -4.29 \\
Mid & +2.05 \\
Common & +16.98 \\
\end{longtable}}
}

\begin{figure}
\centering
\pandocbounded{\includegraphics[width=0.51\linewidth,keepaspectratio,alt={Signed option-level error against human option share, all 206 option rows. The direct readout shows the two failure modes: collapse of mid-prevalence options toward zero selection and saturation of salient options. The probability readout is flatter through the rare and mid range at the cost of a small positive floor on rare options. Hollow markers are the max-2-capped battery; vertical ticks are uncapped batteries.}]{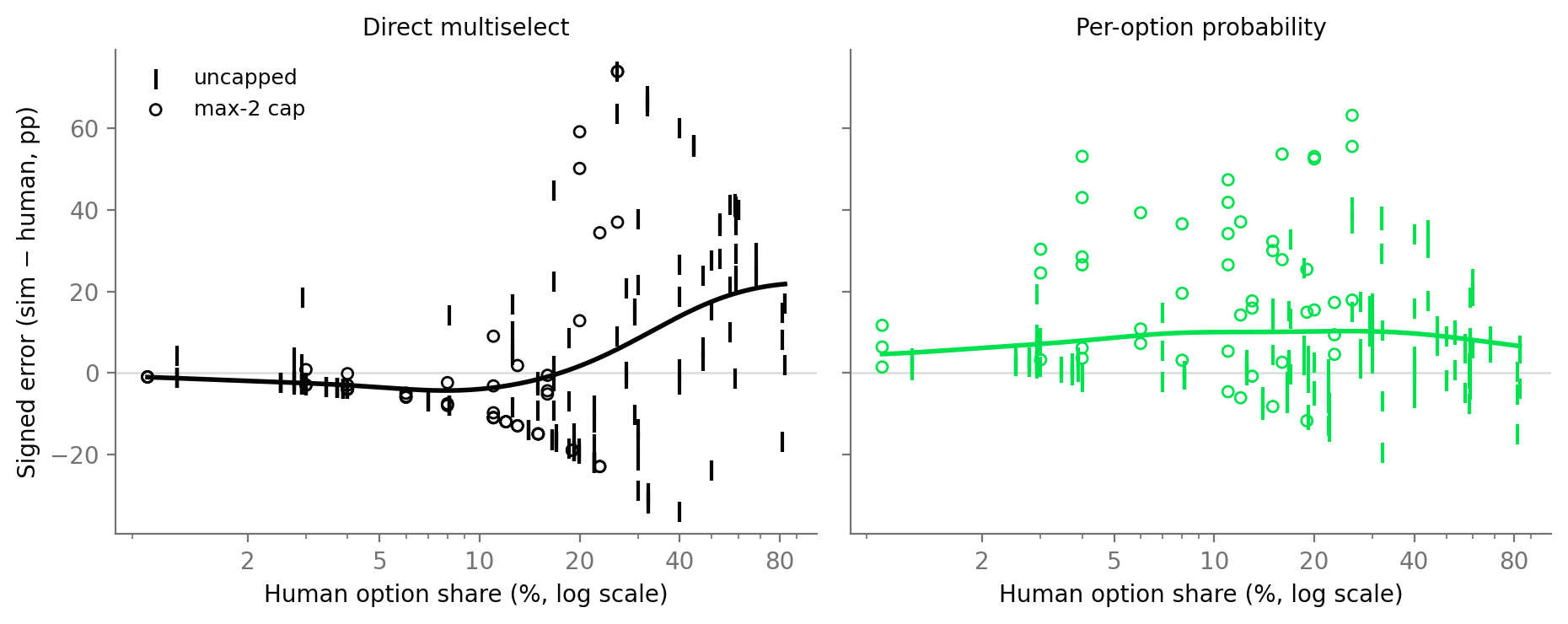}}
\caption{Signed option-level error against human option share, all 206
option rows. The direct readout shows the two failure modes: collapse of
mid-prevalence options toward zero selection and saturation of salient
options. The probability readout is flatter through the rare and mid
range at the cost of a small positive floor on rare options. Hollow
markers are the max-2-capped battery; vertical ticks are uncapped
batteries.}
\end{figure}

Figure 1 plots the option rows. The descriptive pattern is a prevalence
gradient. Where an option is common the committed-set readout often
saturates or collapses it; where an option is rare, collapse is cheap
under absolute error while a positive probability floor is costly. The
direct readout is therefore ahead by 4.29 points on the rare tercile.
Signed errors correlate at \(r=0.44\). These pooled summaries locate a
possible mechanism but do not identify one.

\textbf{No meta-analytic pool.} An earlier analysis applied a
DerSimonian--Laird random-effects formula to six battery means using
between-model spread as if it were sampling variance. That substitution
is not valid: the instruments are not a random sample from a defined
battery population and model disagreement is not a standard error. We
therefore report no pooled effect, \(I^2\), \(\tau^2\), or prediction
interval. \texttt{analysis/pooled\_effects.py} remains in the artifact
as a superseded descriptive calculation and must not be cited
inferentially.

\subsubsection{A.3 Why MAE plus paired bootstrap remains the primary
analysis}\label{a.3-why-mae-plus-paired-bootstrap-remains-the-primary-analysis}

The confirmatory claim is a population-marginal accuracy comparison
between two elicitation contracts on identical respondents; the
option-marginal MAE with a paired bootstrap clustered by distinct prompt
is the estimand-matched test of that claim and mirrors the analysis used
in prior express-versus-simulate comparisons for single-choice items.
The regressions above condition on human prevalence, which is part of
the outcome scale, so they are informative about mechanism but would be
a biased primary endpoint. We therefore report them as explanatory only.

\subsubsection{A.4 Battery-level effects,
plotted}\label{a.4-battery-level-effects-plotted}

Figure 2 plots the Section 5 table: bars are the mean over the three to
five models per battery, ticks the individual models, and the capped
German battery is the only reversal.

\begin{figure}
\centering
\pandocbounded{\includegraphics[width=0.51\linewidth,keepaspectratio,alt={Direct multiselect versus per-option probability elicitation across the six batteries, bars showing the mean over the three to five models per battery and ticks the individual models. Lower is better. The max-2-capped German battery is the only reversal; Appendix A.5 shows that projecting the vectors onto the cap removes it.}]{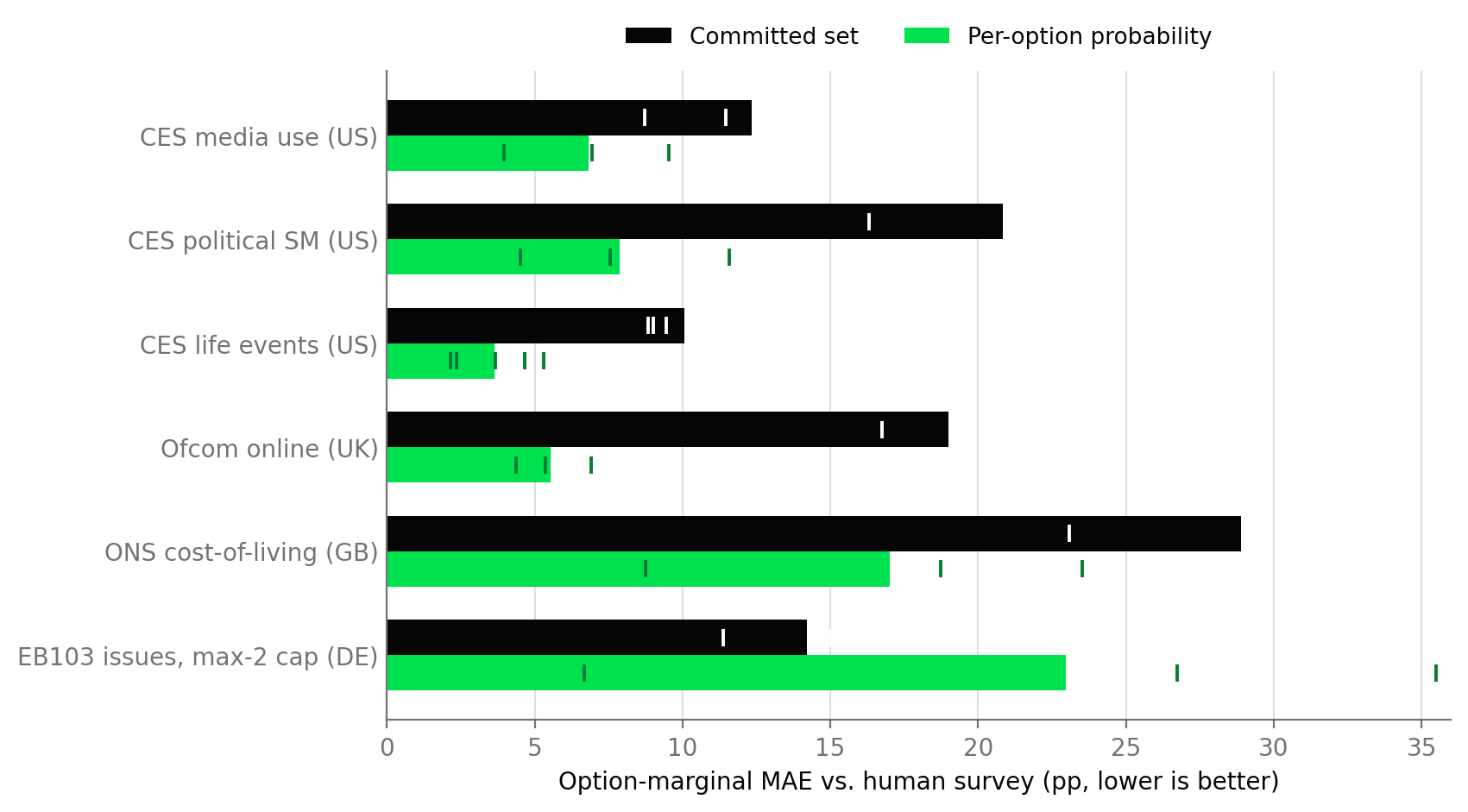}}
\caption{Direct multiselect versus per-option probability elicitation
across the six batteries, bars showing the mean over the three to five
models per battery and ticks the individual models. Lower is better. The
max-2-capped German battery is the only reversal; Appendix A.5 shows
that projecting the vectors onto the cap removes it.}
\end{figure}

\subsubsection{A.5 Constraint projection sensitivity
(exploratory)}\label{a.5-constraint-projection-sensitivity-exploratory}

Section 5 reports the reversal on the one instrument that caps answers
at two and treats the cap as a candidate moderator. The mass gives a
positive test. If the reversal is caused by probabilities ignoring the
constraint, then imposing the constraint should remove it, and the
constraint is a design fact of the instrument rather than target
information, so the test uses no human data.

For each persona we scale the elicited vector by
\(\min(1, C/\sum_o p_o)\) with \(C = 2\), the stated maximum, and
rescore. Scoring and option handling follow
\texttt{score\_allbatteries.py} exactly.

{\def\LTcaptype{none} 
{\footnotesize\begin{longtable}[]{@{}
  >{\raggedright\arraybackslash}p{(\linewidth - 12\tabcolsep) * \real{0.1429}}
  >{\raggedright\arraybackslash}p{(\linewidth - 12\tabcolsep) * \real{0.1429}}
  >{\raggedright\arraybackslash}p{(\linewidth - 12\tabcolsep) * \real{0.1429}}
  >{\raggedright\arraybackslash}p{(\linewidth - 12\tabcolsep) * \real{0.1429}}
  >{\raggedright\arraybackslash}p{(\linewidth - 12\tabcolsep) * \real{0.1429}}
  >{\raggedright\arraybackslash}p{(\linewidth - 12\tabcolsep) * \real{0.1429}}
  >{\raggedright\arraybackslash}p{(\linewidth - 12\tabcolsep) * \real{0.1429}}@{}}
\toprule\noalign{}
\begin{minipage}[b]{\linewidth}\raggedright
Model
\end{minipage} & \begin{minipage}[b]{\linewidth}\raggedright
Committed set
\end{minipage} & \begin{minipage}[b]{\linewidth}\raggedright
Probability, raw
\end{minipage} & \begin{minipage}[b]{\linewidth}\raggedright
Probability, capped
\end{minipage} & \begin{minipage}[b]{\linewidth}\raggedright
Elicited mass
\end{minipage} & \begin{minipage}[b]{\linewidth}\raggedright
\(\Delta\) raw
\end{minipage} & \begin{minipage}[b]{\linewidth}\raggedright
\(\Delta\) capped
\end{minipage} \\
\midrule\noalign{}
\endhead
\bottomrule\noalign{}
\endlastfoot
claude-haiku-4-5 & 16.25 & 26.75 & 3.49 & 619.9 & \(-10.50\) &
\(+12.76\) \\
gpt-4o-mini & 15.03 & 35.48 & 5.23 & 759.7 & \(-20.46\) & \(+9.79\) \\
gemini-3.7-flash & 11.38 & 6.68 & 5.28 & 235.7 & \(+4.70\) &
\(+6.10\) \\
\end{longtable}}
}

The human selected mass on this instrument is 192.0 points. The two
reversing models emit 620 and 760, three and four times the human total,
and the model that does not reverse emits 236. Projecting onto the cap
removes the reversal on all three: the battery goes from favouring
committed sets on two of three models to favouring probabilities on
three of three, and the worst case improves from \(-20.46\) to \(+9.79\)
(the table's rounded entries differ in the last digit). What this
identifies is the \emph{unconstrained mass}, not the cap itself. The
error is explained by vectors that sum far above any admissible total,
and imposing an admissible total removes it; that is consistent with the
cap being the cause but does not establish it, since the projection
imposes the mass and the cap together and cannot tell them apart; a
within-instrument manipulation of the cap alone, which would, has not
been run on these APIs. The defensible claim is that the reversal is a
property of the missing projection rather than of the probability
contract. Exploratory and specified after the reversal was known;
\texttt{analysis/cap\_projection.py} reproduces the table.

\subsubsection{A.6 Post-stratifying the cohorts
(exploratory)}\label{a.6-post-stratifying-the-cohorts-exploratory}

The cohorts are the first 500 rows of demographic files generated by
seeded independent assignment to one-way marginal targets; their one-way
shares deviate from those targets by up to 5.4 points. The contract
comparison is paired within cohort, but the panel-versus-query
comparison is not. Each cohort is therefore raked to its recorded
one-way margins by iterative proportional fitting and rescored; the
query is unchanged by construction. This cannot repair absent
joint-distribution evidence or the Ofcom/ONS population-frame
mismatches.

{\def\LTcaptype{none} 
{\footnotesize\begin{longtable}[]{@{}lll@{}}
\toprule\noalign{}
Panel & Committed set & Probability \\
\midrule\noalign{}
\endhead
\bottomrule\noalign{}
\endlastfoot
Unweighted & 17.56 & 10.64 \\
Post-stratified & 17.71 & 10.60 \\
\end{longtable}}
}

Entries are means of the six battery means (each battery averaged over
its models), which is why the unweighted row differs from the
eighteen-pair means of B.3 (17.65 and 10.79). The three CC24\_300d
panels are raked to full-population margins although their base is
social-media users, for want of a base-specific margin file; their
contribution to the shift is small.

Raking moves the all-instrument descriptive mean by 0.15 points under
committed sets and 0.04 under probabilities. This shows only that
restoring the recorded one-way margins has little effect on that
summary; it does not validate the independent-assignment cohort,
filtered bases, or population frames. \texttt{analysis/poststratify.py}
reproduces the table given the demographic file.

\subsubsection{A.7 Implied set size}\label{a.7-implied-set-size}

The marginal endpoint says nothing about set size, but for a check-all
item the human mean number of substantive options selected equals the
sum of the option marginals, so the released targets determine it, and
each contract implies one: the mean size of the committed sets, and the
sum of the mean per-option probabilities, which is the expected set size
under an independent-Bernoulli reading of the vector. Explicit
none-codes are excluded on both sides.
\texttt{analysis/implied\_cardinality.py} computes the table from the
retained responses.

{\def\LTcaptype{none} 
{\footnotesize\begin{longtable}[]{@{}
  >{\raggedright\arraybackslash}p{(\linewidth - 14\tabcolsep) * \real{0.1250}}
  >{\raggedright\arraybackslash}p{(\linewidth - 14\tabcolsep) * \real{0.1250}}
  >{\raggedright\arraybackslash}p{(\linewidth - 14\tabcolsep) * \real{0.1250}}
  >{\raggedright\arraybackslash}p{(\linewidth - 14\tabcolsep) * \real{0.1250}}
  >{\raggedright\arraybackslash}p{(\linewidth - 14\tabcolsep) * \real{0.1250}}
  >{\raggedright\arraybackslash}p{(\linewidth - 14\tabcolsep) * \real{0.1250}}
  >{\raggedright\arraybackslash}p{(\linewidth - 14\tabcolsep) * \real{0.1250}}
  >{\raggedright\arraybackslash}p{(\linewidth - 14\tabcolsep) * \real{0.1250}}@{}}
\toprule\noalign{}
\begin{minipage}[b]{\linewidth}\raggedright
Battery
\end{minipage} & \begin{minipage}[b]{\linewidth}\raggedright
Model
\end{minipage} & \begin{minipage}[b]{\linewidth}\raggedright
Human mean set size
\end{minipage} & \begin{minipage}[b]{\linewidth}\raggedright
Committed set
\end{minipage} & \begin{minipage}[b]{\linewidth}\raggedright
Probability
\end{minipage} & \begin{minipage}[b]{\linewidth}\raggedright
Committed / human
\end{minipage} & \begin{minipage}[b]{\linewidth}\raggedright
Probability / human
\end{minipage} & \begin{minipage}[b]{\linewidth}\raggedright
Closer
\end{minipage} \\
\midrule\noalign{}
\endhead
\bottomrule\noalign{}
\endlastfoot
CC24\_300 & haiku-4-5 & 1.96 & 2.55 & 2.31 & 1.31 & 1.18 &
probability \\
CC24\_300 & gpt-4o-mini & 1.96 & 2.16 & 1.99 & 1.10 & 1.02 &
probability \\
CC24\_300 & gemini-3.7-flash & 1.96 & 2.33 & 2.05 & 1.19 & 1.05 &
probability \\
CC24\_300d & haiku-4-5 & 1.36 & 1.01 & 1.29 & 0.75 & 0.95 &
probability \\
CC24\_300d & gpt-4o-mini & 1.36 & 0.85 & 1.38 & 0.62 & 1.01 &
probability \\
CC24\_300d & gemini-3.7-flash & 1.36 & 1.21 & 1.14 & 0.89 & 0.84 &
committed set \\
IN5A & haiku-4-5 & 4.30 & 5.19 & 4.62 & 1.21 & 1.07 & probability \\
IN5A & gpt-4o-mini & 4.30 & 5.43 & 4.55 & 1.26 & 1.06 & probability \\
IN5A & gemini-3.7-flash & 4.30 & 6.14 & 4.88 & 1.43 & 1.14 &
probability \\
OPN & haiku-4-5 & 2.74 & 4.15 & 4.74 & 1.51 & 1.73 & committed set \\
OPN & gpt-4o-mini & 2.74 & 5.10 & 5.24 & 1.86 & 1.91 & committed set \\
OPN & gemini-3.7-flash & 2.74 & 4.55 & 3.59 & 1.66 & 1.31 &
probability \\
QA3 (cap 2) & haiku-4-5 & 1.91 & 2.00 & 6.13 & 1.05 & 3.21 & committed
set \\
QA3 (cap 2) & gpt-4o-mini & 1.91 & 2.00 & 7.47 & 1.05 & 3.91 & committed
set \\
QA3 (cap 2) & gemini-3.7-flash & 1.91 & 2.00 & 2.33 & 1.05 & 1.22 &
committed set \\
CC24\_305 & haiku-4-5 & 1.39 & 1.62 & 1.82 & 1.17 & 1.31 & committed
set \\
CC24\_305 & gpt-4o-mini & 1.39 & 2.06 & 1.63 & 1.48 & 1.18 &
probability \\
CC24\_305 & gemini-3.7-flash & 1.39 & 1.85 & 1.79 & 1.33 & 1.29 &
probability \\
CC24\_305 & sonnet-5 & 1.39 & 1.34 & 1.48 & 0.96 & 1.07 & committed
set \\
CC24\_305 & gpt-5.6-terra & 1.39 & 1.10 & 1.47 & 0.79 & 1.06 &
probability \\
\end{longtable}}
}

In the aligned uncapped subset, probabilities are closer to the human
set size in 6 of 8 runs and committed sets in 2. The all-instrument
sensitivity count is 12 of 17 versus 5. On the capped battery the
committed sets sit at the cap while raw vectors imply 1.2 to 3.9 times
the human mean. The probability figure is an expectation under
independence and says nothing about the distribution of realised sets or
co-selection.

\subsubsection{A.8 Realized sets from the vectors
(exploratory)}\label{a.8-realized-sets-from-the-vectors-exploratory}

The set-size comparison of A.7 uses the expected size of the vector.
This section realizes sets from the vectors, by independent Bernoulli
draws per option (200 draws, seed 20260901), and scores them against the
CES microdata on two joint endpoints: the Wasserstein-1 distance between
set-size distributions and the mean absolute error of pairwise
co-selection incidence. The shuffled column permutes each option's
elicited probabilities across personas before realizing, which preserves
every marginal and destroys within-person structure.
\texttt{analysis/realize\_sets\_basemodels.py} and
\texttt{analysis/shuffle\_baseline.py} compute the table given the
microdata.

{\def\LTcaptype{none} 
{\footnotesize\begin{longtable}[]{@{}
  >{\raggedright\arraybackslash}p{(\linewidth - 8\tabcolsep) * \real{0.2000}}
  >{\raggedright\arraybackslash}p{(\linewidth - 8\tabcolsep) * \real{0.2000}}
  >{\raggedright\arraybackslash}p{(\linewidth - 8\tabcolsep) * \real{0.2000}}
  >{\raggedright\arraybackslash}p{(\linewidth - 8\tabcolsep) * \real{0.2000}}
  >{\raggedright\arraybackslash}p{(\linewidth - 8\tabcolsep) * \real{0.2000}}@{}}
\toprule\noalign{}
\begin{minipage}[b]{\linewidth}\raggedright
Battery
\end{minipage} & \begin{minipage}[b]{\linewidth}\raggedright
Model
\end{minipage} & \begin{minipage}[b]{\linewidth}\raggedright
Set-size \(W_1\): committed / realized
\end{minipage} & \begin{minipage}[b]{\linewidth}\raggedright
Pairwise MAE: committed / realized
\end{minipage} & \begin{minipage}[b]{\linewidth}\raggedright
Shuffled: \(W_1\) / pairwise
\end{minipage} \\
\midrule\noalign{}
\endhead
\bottomrule\noalign{}
\endlastfoot
CC24\_305 & haiku-4-5 & 0.41 / 0.57 & 1.06 / 1.26 & 0.58 / 1.31 \\
CC24\_305 & gpt-4o-mini & 0.60 / 0.39 & 1.79 / 1.11 & 0.41 / 1.23 \\
CC24\_305 & sonnet-5 & 0.24 / 0.38 & 1.03 / 0.98 & 0.40 / 1.02 \\
CC24\_305 & gpt-5.6-terra & 0.46 / 0.34 & 0.93 / 0.91 & 0.36 / 0.95 \\
CC24\_305 & gemini-3.7-flash & 0.40 / 0.42 & 1.58 / 0.95 & 0.44 /
0.98 \\
CC24\_300 & haiku-4-5 & 0.68 / 0.40 & 10.31 / 6.77 & 0.41 / 6.59 \\
CC24\_300 & gpt-4o-mini & 0.30 / 0.13 & 6.54 / 3.95 & 0.17 / 2.97 \\
CC24\_300 & gemini-3.7-flash & 0.34 / 0.16 & 5.09 / 2.78 & 0.19 /
2.57 \\
CC24\_300d & haiku-4-5 & 0.67 / 0.38 & 7.65 / 5.88 & 0.38 / 6.01 \\
CC24\_300d & gpt-4o-mini & 0.81 / 0.41 & 7.29 / 5.48 & 0.40 / 5.81 \\
CC24\_300d & gemini-3.7-flash & 0.47 / 0.38 & 6.24 / 7.47 & 0.39 /
7.83 \\
\end{longtable}}
}

Realized sets beat the committed sets on set-size distribution in 8 of
11 combinations and on co-selection in 9 of 11; on CC24\_305 the
realized sets are \emph{worse} on set size on three of five models. The
shuffled control is within 0.04 of the persona-specific realization on
set size on every combination, and on co-selection within 0.11 on
CC24\_305 but by up to 0.99 elsewhere with mixed sign: the shuffle is
better on the three CC24\_300 combinations and worse on the three
CC24\_300d ones. Whatever persona-specific joint structure the realized
vectors carry is therefore small and battery-dependent, and on one
battery a random rearrangement of the same numbers does better. This is
the result that most limits the probability contract: it is the better
population estimator of Section 5 and the better set-size estimator of
A.7 on most runs, and it is not a reliable source of individual-level
joint structure. Exploratory; one survey family; independence is a
modelling choice.

\subsection{Appendix B. Simulator
specification}\label{appendix-b.-simulator-specification}

The reported runs use named language models through provider APIs,
without a commercial simulation platform, persistent persona store, web
grounding or retrieval. A simulated respondent is one system prompt
naming four demographic attributes, followed by a battery message. The
release stores terminal model-output text where retained, parsed
responses, validity flags, requested parameters and run identifiers. It
does not contain complete provider request/response envelopes, and token
usage or API-resolved model metadata is absent for some calls. The
included artifacts support recomputing the reported scores from the
retained terminal responses; they do not constitute full transport-level
provenance.

\subsubsection{B.1 Base-model replication
specification}\label{b.1-base-model-replication-specification}

Respondents: the first 500 rows of the cohort-B demographic file
(gender, age band, Census region, education), generated to recorded
one-way margins with seeded independent assignment. Each row becomes a
system prompt describing one United States adult with exactly those
attributes. Question and options are as stored. Direct contract: one
completion at requested temperature 1.0 where accepted, restricted to a
JSON array of option strings, with up to two re-asks on invalid output.
Probability contract: one completion at requested temperature 0.2 where
accepted, mapping each option to an integer 0--100. Concurrency 16; 400
output-token cap, raised to 2,000 for one model after truncation.

{\def\LTcaptype{none} 
{\footnotesize\begin{longtable}[]{@{}
  >{\raggedright\arraybackslash}p{(\linewidth - 12\tabcolsep) * \real{0.1429}}
  >{\raggedright\arraybackslash}p{(\linewidth - 12\tabcolsep) * \real{0.1429}}
  >{\raggedright\arraybackslash}p{(\linewidth - 12\tabcolsep) * \real{0.1429}}
  >{\raggedright\arraybackslash}p{(\linewidth - 12\tabcolsep) * \real{0.1429}}
  >{\raggedright\arraybackslash}p{(\linewidth - 12\tabcolsep) * \real{0.1429}}
  >{\raggedright\arraybackslash}p{(\linewidth - 12\tabcolsep) * \real{0.1429}}
  >{\raggedright\arraybackslash}p{(\linewidth - 12\tabcolsep) * \real{0.1429}}@{}}
\toprule\noalign{}
\begin{minipage}[b]{\linewidth}\raggedright
Model
\end{minipage} & \begin{minipage}[b]{\linewidth}\raggedright
n direct
\end{minipage} & \begin{minipage}[b]{\linewidth}\raggedright
n prob
\end{minipage} & \begin{minipage}[b]{\linewidth}\raggedright
Direct MAE
\end{minipage} & \begin{minipage}[b]{\linewidth}\raggedright
Probability MAE
\end{minipage} & \begin{minipage}[b]{\linewidth}\raggedright
Difference (95\% CI, clustered)
\end{minipage} & \begin{minipage}[b]{\linewidth}\raggedright
Reduction
\end{minipage} \\
\midrule\noalign{}
\endhead
\bottomrule\noalign{}
\endlastfoot
claude-sonnet-5 & 500 & 500 & 8.998 & 2.357 & +6.64 {[}6.18, 7.40{]} &
74\% \\
gpt-5.6-terra & 500 & 500 & 9.452 & 2.149 & +7.30 {[}6.65, 7.75{]} &
77\% \\
gemini-3.7-flash & 500 & 500 & 11.807 & 4.661 & +7.15 {[}6.59, 7.77{]} &
61\% \\
claude-haiku-4-5-20251001 & 500 & 500 & 8.838 & 3.659 & +5.18 {[}4.41,
6.24{]} & 59\% \\
gpt-4o-mini & 498 & 500 & 11.229 & 5.305 & +5.92 {[}4.93, 6.83{]} &
53\% \\
\end{longtable}}
}

Intervals are paired bootstraps clustered by distinct prompt (120
clusters); the respondent-bootstrap intervals, which are narrower, are
+6.64 {[}6.33, 7.02{]}, +7.30 {[}6.97, 7.60{]}, +7.15 {[}6.81, 7.47{]},
+5.18 {[}4.73, 5.68{]} and +5.92 {[}5.29, 6.49{]}.

Requested and accepted decoding settings are recorded at run level where
available: the older pair accepted temperature 1.0 (direct) / 0.2
(probability); claude-sonnet-5 rejected 0.2 on probability;
gpt-5.6-terra rejected non-default temperature; and gemini-3.7-flash
accepted both settings. claude-haiku-4-5-20251001 is a dated identifier.
The gpt-4o-mini dated resolution and the remaining alias-resolution
notes were assembled after collection rather than retained in every
provider envelope, so they are provenance annotations, not cryptographic
model pins. The two invalid gpt-4o-mini CC24\_305 direct responses were
persistent empty arrays after three asks and were excluded pairwise. The
artifact records an estimated five-model cost below five US dollars
using documented pricing assumptions, but lacks complete per-call usage
envelopes or an invoice reconciliation.

\subsubsection{B.2 Contract by
temperature}\label{b.2-contract-by-temperature}

Contract and decoding temperature are separable only if both contracts
run at the same temperature. Of the five models, claude-sonnet-5 and
gpt-5.6-terra reject a non-default temperature argument outright
(verbatim refusals archived in the release package), so both of their
contracts ran at the model default and were already matched. The
remaining three support a full contract-by-temperature design, 500
respondents per cell, all cells validated with zero invalid records:

{\def\LTcaptype{none} 
{\footnotesize\begin{longtable}[]{@{}
  >{\raggedright\arraybackslash}p{(\linewidth - 14\tabcolsep) * \real{0.1250}}
  >{\raggedright\arraybackslash}p{(\linewidth - 14\tabcolsep) * \real{0.1250}}
  >{\raggedright\arraybackslash}p{(\linewidth - 14\tabcolsep) * \real{0.1250}}
  >{\raggedright\arraybackslash}p{(\linewidth - 14\tabcolsep) * \real{0.1250}}
  >{\raggedright\arraybackslash}p{(\linewidth - 14\tabcolsep) * \real{0.1250}}
  >{\raggedright\arraybackslash}p{(\linewidth - 14\tabcolsep) * \real{0.1250}}
  >{\raggedright\arraybackslash}p{(\linewidth - 14\tabcolsep) * \real{0.1250}}
  >{\raggedright\arraybackslash}p{(\linewidth - 14\tabcolsep) * \real{0.1250}}@{}}
\toprule\noalign{}
\begin{minipage}[b]{\linewidth}\raggedright
Model
\end{minipage} & \begin{minipage}[b]{\linewidth}\raggedright
direct@T0.2
\end{minipage} & \begin{minipage}[b]{\linewidth}\raggedright
direct@T1.0
\end{minipage} & \begin{minipage}[b]{\linewidth}\raggedright
prob@T0.2
\end{minipage} & \begin{minipage}[b]{\linewidth}\raggedright
prob@T1.0
\end{minipage} & \begin{minipage}[b]{\linewidth}\raggedright
readout effect @T0.2
\end{minipage} & \begin{minipage}[b]{\linewidth}\raggedright
readout effect @T1.0
\end{minipage} & \begin{minipage}[b]{\linewidth}\raggedright
temp. effect (direct)
\end{minipage} \\
\midrule\noalign{}
\endhead
\bottomrule\noalign{}
\endlastfoot
claude-haiku-4-5-20251001 & 9.053 & 8.838 & 3.659 & 3.645 & +5.39
{[}4.96, 5.83{]} & +5.19 {[}4.72, 5.68{]} & -0.215, CI crosses zero \\
gpt-4o-mini & 11.659 & 11.229 & 5.305 & 5.317 & +6.35 {[}5.85, 6.79{]} &
+5.92 {[}5.31, 6.49{]} & -0.479, CI crosses zero \\
gemini-3.7-flash & 11.776 & 11.807 & 4.661 & 4.630 & +7.12 {[}6.77,
7.46{]} & +7.18 {[}6.84, 7.50{]} & +0.031, CI crosses zero \\
\end{longtable}}
}

Table entries are full-sample MAEs; the paired estimates and intervals
use the 498 or 500 personas valid in both cells, so a row's cells and
its paired difference can differ in the second decimal. Temperature main
effects within a contract (1.0 minus 0.2) are -0.21 {[}-0.54, 0.15{]},
-0.48 {[}-1.05, 0.11{]}, and +0.03 {[}-0.26, 0.32{]} for the direct
contract and -0.01, +0.01, and -0.03 for the probability contract, every
interval covering zero. Contract-by-temperature interactions are -0.20,
-0.44, and +0.06.

\textbf{Test-retest.} We re-collected CC24\_305 on claude-haiku-4-5 end
to end under both contracts, with cohort, prompts and parameters
identical to the confirmatory run, and separately found a second
collection of the committed-set cell at temperature 0.2 inside the grid
above.

{\def\LTcaptype{none} 
{\footnotesize\begin{longtable}[]{@{}llllll@{}}
\toprule\noalign{}
& Run 1 & Run 2 & Change & Per-option mean \textbar{} max & Zero
coverage \\
\midrule\noalign{}
\endhead
\bottomrule\noalign{}
\endlastfoot
Committed set & 8.838 & 9.870 & 1.03 & 2.57 \textbar{} 23.20 & 9 to 8 \\
Probability & 3.659 & 4.627 & 0.97 & 1.48 \textbar{} 5.57 & 0 to 0 \\
Readout effect & \(+5.179\) & \(+5.243\) & \textbf{0.064} & & \\
Committed set @0.2 & 9.053 & 10.101 & 1.05 & 3.08 \textbar{} 25.80 & \\
\end{longtable}}
}

Three things follow, and the third is the one that matters for reading
this paper. First, the committed-set contract is not merely less
accurate but markedly less reliable: between collections its per-option
shares move by 2.57 points on average and by 23.2 at the extreme,
against 1.48 and 5.57 for the probability readout. Second, the coverage
asymmetry replicates rather than being an artifact of one collection:
the probability readout returns no zeros in either run and the
committed-set readout returns nine and eight. Third, and correcting a
natural inference from the first two, absolute MAE is the unstable
quantity and the paired contrast is not. Levels move by about a point
between collections, but the readout effect itself moves by 0.064,
roughly one sixteenth of that. The confirmatory effect estimates in this
paper are paired within-cohort differences, which are the quantities
that replicate; the absolute MAE levels are not stable to a point and
should not be read as fidelity scores. Temperature main effects, at most
0.48, are smaller than the level noise and should be treated as
indistinguishable from it. \texttt{analysis/test\_retest.py} reproduces
the table.

\textbf{Matched-temperature replication.} The main comparison runs each
contract at its conventional decoding setting. To remove temperature as
a factor entirely we reran the committed-set contract at 0.2, matching
the probability contract, on claude-haiku-4-5 across five batteries.

{\def\LTcaptype{none} 
{\footnotesize\begin{longtable}[]{@{}
  >{\raggedright\arraybackslash}p{(\linewidth - 10\tabcolsep) * \real{0.1667}}
  >{\raggedright\arraybackslash}p{(\linewidth - 10\tabcolsep) * \real{0.1667}}
  >{\raggedright\arraybackslash}p{(\linewidth - 10\tabcolsep) * \real{0.1667}}
  >{\raggedright\arraybackslash}p{(\linewidth - 10\tabcolsep) * \real{0.1667}}
  >{\raggedright\arraybackslash}p{(\linewidth - 10\tabcolsep) * \real{0.1667}}
  >{\raggedright\arraybackslash}p{(\linewidth - 10\tabcolsep) * \real{0.1667}}@{}}
\toprule\noalign{}
\begin{minipage}[b]{\linewidth}\raggedright
Battery
\end{minipage} & \begin{minipage}[b]{\linewidth}\raggedright
Committed set @1.0
\end{minipage} & \begin{minipage}[b]{\linewidth}\raggedright
Committed set @0.2
\end{minipage} & \begin{minipage}[b]{\linewidth}\raggedright
Probability @0.2
\end{minipage} & \begin{minipage}[b]{\linewidth}\raggedright
Difference @1.0
\end{minipage} & \begin{minipage}[b]{\linewidth}\raggedright
Difference matched
\end{minipage} \\
\midrule\noalign{}
\endhead
\bottomrule\noalign{}
\endlastfoot
A CC24 300 & 16.80 & 16.56 & 9.55 & +7.26 & +7.02 \\
B CC24 305 & 8.84 & 10.10 & 3.66 & +5.18 & +6.44 \\
C IN5A & 19.38 & 18.71 & 5.37 & +14.01 & +13.34 \\
D OPN & 29.62 & 30.00 & 18.73 & +10.89 & +11.27 \\
E QA3 & 16.25 & 16.23 & 26.75 & -10.50 & -10.52 \\
\end{longtable}}
}

Matching temperature moves the contract difference by +0.18 points on
average across the uncapped batteries: all four uncapped batteries still
favour the probability contract, from \(+6.44\) to \(+13.34\), and the
capped battery still reverses at \(-10.52\) against \(-10.50\). The
CC24\_305 row uses the second end-to-end collection of the committed-set
cell at 0.2 (10.10); with the collection from the grid above (9.05) the
row's difference is \(+5.39\) and the four-battery mean shift is
\(-0.08\). Neither choice changes a sign;
\texttt{matched\_temp/score\_matched\_temp.py} reports both. Decoding
temperature is not a competing explanation for any result in this paper,
though that is because the contract effect is large rather than because
temperature has been estimated precisely.

\subsubsection{B.3 Population-prevalence
baseline}\label{b.3-population-prevalence-baseline}

No personas. For each battery, one prompt asks the model to estimate the
percentage of the stated target population selecting each option, with a
neutral system prompt and the released question and option text; twenty
independent repetitions per battery per model, averaged. Option-marginal
MAE against the same target files:

{\def\LTcaptype{none} 
{\footnotesize\begin{longtable}[]{@{}
  >{\raggedright\arraybackslash}p{(\linewidth - 8\tabcolsep) * \real{0.2000}}
  >{\raggedright\arraybackslash}p{(\linewidth - 8\tabcolsep) * \real{0.2000}}
  >{\raggedright\arraybackslash}p{(\linewidth - 8\tabcolsep) * \real{0.2000}}
  >{\raggedright\arraybackslash}p{(\linewidth - 8\tabcolsep) * \real{0.2000}}
  >{\raggedright\arraybackslash}p{(\linewidth - 8\tabcolsep) * \real{0.2000}}@{}}
\toprule\noalign{}
\begin{minipage}[b]{\linewidth}\raggedright
Battery
\end{minipage} & \begin{minipage}[b]{\linewidth}\raggedright
claude-sonnet-5
\end{minipage} & \begin{minipage}[b]{\linewidth}\raggedright
gpt-5.6-terra
\end{minipage} & \begin{minipage}[b]{\linewidth}\raggedright
claude-haiku-4-5
\end{minipage} & \begin{minipage}[b]{\linewidth}\raggedright
3-model mean
\end{minipage} \\
\midrule\noalign{}
\endhead
\bottomrule\noalign{}
\endlastfoot
CC24\_300 & 6.60 & 15.80 & 13.36 & 11.92 \\
CC24\_300d & 12.28 & 20.27 & 15.10 & 15.88 \\
CC24\_305 & 2.56 & 2.13 & 2.90 & 2.53 \\
IN5A & 8.55 & 6.14 & 8.31 & 7.67 \\
OPN & 8.77 & 4.70 & 7.60 & 7.02 \\
QA3 (capped) & 7.15 & 4.00 & 5.71 & 5.62 \\
\end{longtable}}
}

All 360 calls returned valid structured estimates
(\path{base_model_replication/responses_population_{sonnet5,gpt56terra,haiku45}.jsonl}).
The mean absolute difference between the twenty-run average and the mean
single-run MAE is 0.20 points (maximum 0.80), with per-option across-run
standard deviations of 0.7 to 3.6 points: the residual error is bias,
not run-to-run variance, so twenty queries are not materially better
than one. On CC24\_305 the baseline (2.13 to 2.90) sits inside the range
of the demographic-only 500-persona probability contract of Appendix B.1
(2.15 to 5.31). The three models in this table were chosen for the
contamination probes of Appendix B.4, and only claude-haiku-4-5 among
them also ran as a panel. \textbf{Panel comparison, model-matched.}
Comparing a query on one set of models with a panel on another would
confound the response contract with model quality, so the query was also
run on exactly the models each panel was run on. Every row below is one
model asked both ways: once as a single population-prevalence query with
no personas, once as a panel of up to 500 personas under each contract,
scored against the same released target.

{\def\LTcaptype{none} 
{\footnotesize\begin{longtable}[]{@{}
  >{\raggedright\arraybackslash}p{(\linewidth - 10\tabcolsep) * \real{0.1667}}
  >{\raggedright\arraybackslash}p{(\linewidth - 10\tabcolsep) * \real{0.1667}}
  >{\raggedright\arraybackslash}p{(\linewidth - 10\tabcolsep) * \real{0.1667}}
  >{\raggedright\arraybackslash}p{(\linewidth - 10\tabcolsep) * \real{0.1667}}
  >{\raggedright\arraybackslash}p{(\linewidth - 10\tabcolsep) * \real{0.1667}}
  >{\raggedright\arraybackslash}p{(\linewidth - 10\tabcolsep) * \real{0.1667}}@{}}
\toprule\noalign{}
\begin{minipage}[b]{\linewidth}\raggedright
Battery
\end{minipage} & \begin{minipage}[b]{\linewidth}\raggedright
Model
\end{minipage} & \begin{minipage}[b]{\linewidth}\raggedright
Query, 20 reps
\end{minipage} & \begin{minipage}[b]{\linewidth}\raggedright
Query, single rep
\end{minipage} & \begin{minipage}[b]{\linewidth}\raggedright
Panel committed
\end{minipage} & \begin{minipage}[b]{\linewidth}\raggedright
Panel probability
\end{minipage} \\
\midrule\noalign{}
\endhead
\bottomrule\noalign{}
\endlastfoot
CC24\_300 & haiku45 / gpt4omini / gemini37flash & 8.89 / 10.90 / 4.87 &
8.98 / 10.97 / 5.07 & 16.80 / 11.46 / 8.72 & 9.55 / 6.93 / 3.97 \\
CC24\_300d & haiku45 / gpt4omini / gemini37flash & 6.29 / 9.74 / 11.79 &
6.54 / 10.53 / 11.98 & 24.91 / 16.30 / 21.33 & 4.52 / 11.59 / 7.54 \\
CC24\_305 & haiku45 / gpt4omini / gemini37flash & 3.16 / 4.94 / 4.31 &
3.36 / 5.36 / 4.33 & 8.84 / 11.23 / 11.81 & 3.66 / 5.31 / 4.66 \\
IN5A & haiku45 / gpt4omini / gemini37flash & 6.43 / 8.29 / 2.52 & 6.85 /
8.53 / 3.40 & 19.38 / 16.76 / 20.89 & 5.37 / 4.38 / 6.90 \\
OPN & haiku45 / gpt4omini / gemini37flash & 7.68 / 7.71 / 5.57 & 8.25 /
8.43 / 5.64 & 29.62 / 33.96 / 23.07 & 18.73 / 23.53 / 8.74 \\
QA3 & haiku45 / gpt4omini / gemini37flash & 8.44 / 5.84 / 5.04 & 8.62 /
6.27 / 5.10 & 16.25 / 15.03 / 11.38 & 26.75 / 35.48 / 6.68 \\
\textbf{aligned mean, 9 pairs} & CC24\_300, CC24\_305, QA3 &
\textbf{6.27} & 6.45 & 12.39 & 11.44 \\
\textbf{all-instrument sensitivity mean, 18 pairs} & & 6.80 & 7.12 &
17.65 & 10.79 \\
\end{longtable}}
}

Across the nine population-aligned pairs, the averaged query beats
committed sets on \textbf{9 of 9} and raw probability vectors on 7 of 9;
a single repetition averages 6.45 and beats them on 9 of 9 and 6 of 9.
After applying the QA3 constraint projection, probability vectors
average 5.34 and the query beats them on 5 of 9. Thus the query
diagnoses the weakness of committed-panel marginal validation; it is not
the best estimator on average. The all-instrument sensitivity summary is
18 of 18 versus committed sets and 12 of 18 versus raw vectors, but it
includes CC24\_300d's endogenous base and the IN5A/ONS frame mismatches.
Three query cells returned fewer than twenty valid estimates and are
reported as scored; the retained repetitions are in
\texttt{analysis/prevalence\_matched\_raw.jsonl}.

The query nonetheless produces no sets, no cardinality, no co-selection
and no individual, and Section 6 shows its advantage does not transfer
to targets with no published table.
\texttt{analysis/prevalence\_matched.py} reproduces the table.

\subsubsection{B.4 Contamination probe}\label{b.4-contamination-probe}

Whether the prevalence baseline estimates or retrieves matters for every
result in this paper, because the same models answer as personas. Two
probe families, 510 calls across three models: probe 1 is 180 calls
(three models, six batteries, ten repetitions), probe 1b is 240 (two
complying models, six batteries, twenty repetitions), and probe 2 is 90
(three models, six batteries, five repetitions). Probe 1 named the
survey, wave and item and offered an explicit ``unknown'' escape: all
180 calls declined, stating they could not reliably recall the published
figures. Probe 1b removed the escape and forced a best recollection:
mean MAE 7.11 and 7.27 for the two complying models against 7.65 and
8.84 for the neutral prompt. Forcing recall therefore \emph{improves}
accuracy, by 0.54 and 1.57 points, the latter an 18 percent reduction,
and we read that as a weak retrieval component we cannot rule out rather
than as evidence of none. It is far smaller than reproduction of a
published table would produce, and inspection shows the named-source
condition reproducing the neutral condition's systematic errors rather
than correcting them (victim of a crime: 3.9 human, 8.4 neutral, 10.1
named), so literal lookup is excluded while a retrieved-and-adjusted
prior is not. Probe 2 asked for spontaneous source attribution. Counting
a reply as correct when it names the fielding organisation or instrument
(Ofcom; the ONS or its Opinions and Lifestyle Survey; Eurobarometer; the
Cooperative Election Study), the correct source was named in 30 of 90
replies (7 of 15 for IN5A, 13 of 15 for OPN, 10 of 15 for QA3) and never
for the three CES batteries, where 20 of 45 replies attributed the
estimate to Pew Research Center and the rest said it rested on no
specific survey. For calibration, the best knowledge-free constant
predictor scores 9.37 on CC24\_305 against the models' 2.13 to 2.90. A
hybrid account, retrieving a national prior and adjusting, is not
excluded by these probes; Appendix I tests it directly on targets with
no published table.

\textbf{Why this is not a temporal control.} The response files label
gpt-4o-mini, and a later provenance note maps that alias to a dated
snapshot, but the complete provider envelope resolving the alias was not
retained contemporaneously for each call. Training-cutoff comparisons
are therefore post hoc metadata checks, not experimental controls. We do
not infer that a model could not have encountered a target, and all
underlying items or earlier waves are public. The only defensible
contamination conclusion is that the probe did not elicit literal
tables; approximate, retrieved or learned priors remain possible.

\subsubsection{B.5 Validity and pairing}\label{b.5-validity-and-pairing}

Of 20,000 requested confirmatory responses, 19,510 were collected; the
490 not collected are the CC24\_300d personas whose CC24\_300 committed
set did not include social media, who were not asked the follow-up
(bases 449, 324 and 482). Of the 19,510, 19,410 were valid on the first
or a re-asked attempt, 99.5 percent, and no committed-set answer on the
capped battery exceeded two options. Thirteen of the twenty runs are
complete at 500 valid responses under both contracts, and the three
CC24\_300d runs are complete at their eligible bases of 449, 324 and
482. The four remaining runs are gpt-4o-mini on CC24\_305 (498 valid
committed sets, two persistent empty arrays), gemini-3.7-flash on IN5A
and on OPN (498 valid probability vectors each, JSON truncated at the
output cap on two calls apiece), and gemini-3.7-flash on the capped QA3
battery, where 94 of 500 probability calls returned malformed JSON after
three attempts and 406 are scored. That last run is the one model that
does not reverse on the capped battery, so its \(+4.70\) is computed on
406 vectors against 500 committed sets; on the 406 personas valid under
both contracts the committed-set error is 11.15 rather than 11.38 and
the difference \(+4.47\) \([3.94, 6.03]\) (clustered), so the direction
is unchanged. Point estimates throughout use every valid record under
each contract; paired bootstrap intervals are computed on the personas
valid under both contracts, matched by respondent identifier, which is
the method of \texttt{base\_model\_replication/score\_all.py},
\texttt{score\_allbatteries.py} and
\texttt{analysis/clustered\_bootstrap.py}.

\subsubsection{B.6 Prompt ablation: is collapse induced by the persona
instruction?}\label{b.6-prompt-ablation-is-collapse-induced-by-the-persona-instruction}

Every persona prompt in this paper ends with the instruction to answer
``based on what is typical and plausible for someone with these
characteristics'' (Section 3). That clause asks for a modal answer, so
zero coverage of a rare option could be the prompt's doing rather than
the contract's. We re-collected CC24\_305 on claude-haiku-4-5 with the
same 500 personas, battery and decoding under three system prompts: the
confirmatory prompt, the same prompt with the clause removed, and a
prompt that names the persona as one person drawn at random from all
adults with those attributes and says that such people differ.
\texttt{base\_model\_replication/runner\_prompt\_ablation.py} collects
it and \texttt{score\_prompt\_ablation.py} scores it; every call was
valid.

{\def\LTcaptype{none} 
{\footnotesize\begin{longtable}[]{@{}
  >{\raggedright\arraybackslash}p{(\linewidth - 16\tabcolsep) * \real{0.1111}}
  >{\raggedright\arraybackslash}p{(\linewidth - 16\tabcolsep) * \real{0.1111}}
  >{\raggedright\arraybackslash}p{(\linewidth - 16\tabcolsep) * \real{0.1111}}
  >{\raggedright\arraybackslash}p{(\linewidth - 16\tabcolsep) * \real{0.1111}}
  >{\raggedright\arraybackslash}p{(\linewidth - 16\tabcolsep) * \real{0.1111}}
  >{\raggedright\arraybackslash}p{(\linewidth - 16\tabcolsep) * \real{0.1111}}
  >{\raggedright\arraybackslash}p{(\linewidth - 16\tabcolsep) * \real{0.1111}}
  >{\raggedright\arraybackslash}p{(\linewidth - 16\tabcolsep) * \real{0.1111}}
  >{\raggedright\arraybackslash}p{(\linewidth - 16\tabcolsep) * \real{0.1111}}@{}}
\toprule\noalign{}
\begin{minipage}[b]{\linewidth}\raggedright
System prompt
\end{minipage} & \begin{minipage}[b]{\linewidth}\raggedright
Committed set MAE
\end{minipage} & \begin{minipage}[b]{\linewidth}\raggedright
Zero-cov.
\end{minipage} & \begin{minipage}[b]{\linewidth}\raggedright
Degenerate
\end{minipage} & \begin{minipage}[b]{\linewidth}\raggedright
Unexpressed (pp)
\end{minipage} & \begin{minipage}[b]{\linewidth}\raggedright
Modal-set share
\end{minipage} & \begin{minipage}[b]{\linewidth}\raggedright
Probability MAE
\end{minipage} & \begin{minipage}[b]{\linewidth}\raggedright
Zero-cov.
\end{minipage} & \begin{minipage}[b]{\linewidth}\raggedright
Readout effect
\end{minipage} \\
\midrule\noalign{}
\endhead
\bottomrule\noalign{}
\endlastfoot
Confirmatory prompt, re-collected & 9.35 & 8 & 9 & 44.6 & 0.35 & 4.59 &
0 & \(+4.76\) \\
Clause removed & 9.01 & 8 & 9 & 44.6 & 0.58 & 5.70 & 0 & \(+3.31\) \\
One randomly drawn person & 7.80 & 8 & 9 & 44.4 & 0.41 & 4.64 & 0 &
\(+3.15\) \\
\end{longtable}}
}

The confirmatory run itself reported 9 zero-coverage options and 8.84 on
this cell (Appendix B.1); the re-collection reproduces it within the
run-to-run variation of Appendix B.2. Removing the clause changes
neither the zero-coverage count (8 / 8 / 8 across the three prompts; the
identity of the zeroed options shifts by one) nor the unexpressed
prevalence, and the random-draw wording, which explicitly asks for
within-cell variation, does not restore the missing options either,
though it lowers the committed-set error. The share of respondents
returning the single most common set \emph{rises} without the clause
(0.35, 0.58, 0.41 across the three prompts), so the clause is not what
makes the sets alike. The readout effect survives under every prompt but
shrinks, from \(+4.76\) to \(+3.31\) and \(+3.15\). Under clause removal
this is mostly because the probability readout worsens; under the
random-draw prompt it is mostly because the committed-set readout
improves. Part of the headline contrast on this battery is therefore
prompt-sensitive. Support collapse is a property of the committed-set
contract jointly with four-attribute conditioning, not of the typicality
instruction; richer grounding reduces it without removing it (Appendix
J.3). One model, one battery; not run on the capped instrument.

\subsection{Appendix C. Clustering the bootstrap by distinct
prompt}\label{appendix-c.-clustering-the-bootstrap-by-distinct-prompt}

A persona is four demographic attributes, so 500 respondents realise
only 57 to 120 distinct prompts and repeated draws from one prompt are
close to deterministic; resampling respondents treats correlated units
as independent. The national-battery intervals in Sections 4 and 5
therefore resample the \emph{distinct prompt} with replacement, carrying
all of that cell's respondents with it. This appendix compares them with
the conventional respondent bootstrap, which the synthetic-panel
literature reports and which is the narrower of the two. The fixed-cell
held-out intervals of Section 6 are a separate respondent-level analysis
of repeated calls under one prompt (Appendix I). 2,000 replicates, seed
20260901.

{\def\LTcaptype{none} 
{\footnotesize\begin{longtable}[]{@{}
  >{\raggedright\arraybackslash}p{(\linewidth - 6\tabcolsep) * \real{0.2500}}
  >{\raggedright\arraybackslash}p{(\linewidth - 6\tabcolsep) * \real{0.2500}}
  >{\raggedright\arraybackslash}p{(\linewidth - 6\tabcolsep) * \real{0.2500}}
  >{\raggedright\arraybackslash}p{(\linewidth - 6\tabcolsep) * \real{0.2500}}@{}}
\toprule\noalign{}
\begin{minipage}[b]{\linewidth}\raggedright
Battery
\end{minipage} & \begin{minipage}[b]{\linewidth}\raggedright
Clusters
\end{minipage} & \begin{minipage}[b]{\linewidth}\raggedright
Width inflation
\end{minipage} & \begin{minipage}[b]{\linewidth}\raggedright
Clustered intervals excluding zero
\end{minipage} \\
\midrule\noalign{}
\endhead
\bottomrule\noalign{}
\endlastfoot
CC24\_300 & 118 & 2.11 to 2.39 & 3 of 3 \\
CC24\_300d & 82 to 114 & 1.62 to 2.29 & 3 of 3 \\
CC24\_305 & 120 & 1.65 to 1.95 & 5 of 5 \\
IN5A & 68 & 2.81 to 3.50 & 3 of 3 \\
OPN & 68 & 2.87 to 3.12 & 3 of 3 \\
QA3 & 56 to 57 & 2.24 to 2.71 & 3 of 3 \\
\end{longtable}}
}

All twenty clustered intervals exclude zero, and the median interval is
2.27 times wider, close to the \(\sqrt{500/118} \approx 2\) that a fully
deterministic within-cell response would imply. The design effect is
instrument-dependent and that is informative: on CC24\_305 clustering
widens the intervals by 1.65 to 1.95, the smallest inflation of the six
batteries, because personas sharing four attributes still answer
somewhat differently on life events, while on the United Kingdom and
German batteries it widens them threefold, because they do not. None of
the paper's conclusions differ between the two; the clustered intervals
are the ones reported in the main text and the respondent bootstrap is
given here for comparability with the literature.
\texttt{analysis/clustered\_bootstrap.py} reproduces the table.

\subsection{Appendix D. The committed-set sampling
floor}\label{appendix-d.-the-committed-set-sampling-floor}

Binary draws carry irreducible binomial noise that probability averaging
does not. Even if every persona drew from the correct inclusion
probability \(q_o/100\), the direct estimator's expected absolute error
per option would be approximately the half-normal mean,

\[\mathbb{E}\left|\hat{q}^{\mathrm{dir}}_{o} - q_o\right| \;\approx\; 100\sqrt{\frac{2}{\pi n}}\sqrt{\frac{q_o}{100}\left(1-\frac{q_o}{100}\right)},\]

about 1.8 percentage points at \(q_o = 50\) and \(n = 500\). This
analytic floor assumes independent Bernoulli draws across 500
respondents; at the effective \(n\) of 120, 68 and 57 distinct prompts
it is 3.7, 4.8 and 5.3 points at 50 percent prevalence. At the sealed
prevalences and effective \(n\) the mean per-option floor is 2.9
(CC24\_300), 3.3 (CC24\_300d), 1.9 (CC24\_305), 4.3 (IN5A), 3.8 (OPN)
and 3.2 (QA3) points. The smallest uncapped readout effect, 4.53 on
CC24\_300, exceeds that battery's floor, and the clustered intervals of
Appendix C carry the effective-\(n\) correction explicitly; a
floor-corrected difference would shrink the CC24\_300 effects to roughly
1.6 to 4.4 points and leave the others largely unchanged.

\subsection{Appendix E. Psychological
account}\label{appendix-e.-psychological-account}

The observed contract sensitivity is consistent with, but does not
prove, established findings in judgment and survey psychology:
preferences are often constructed during elicitation rather than
retrieved (Lichtenstein and Slovic, 1971; Slovic, 1995), and the
compatibility principle predicts that response mode changes which
attributes receive weight (Tversky, Sattath, and Slovic, 1988), so
direct choice and numerical assessment can express different preferences
over unchanged options. A direct multiselect contract compresses the
survey response process (Tourangeau, Rips, and Rasinski, 2000; Schwarz,
1999) into a binary threshold per option; an option-probability contract
lets graded propensity survive aggregation, consistent with the
literature on eliciting subjective expectations (Manski, 2004). The
account has counterpredictions: repeated 0-100 judgments impose more
work and may induce rounding, anchoring, or satisficing (Krosnick,
1991), and independent option probabilities encode neither co-selection
nor cardinality. The studies manipulate synthetic contracts, not human
ones, so no psychological mechanism is identified as causal.

\subsection{Appendix F. Earlier platform-based
studies}\label{appendix-f.-earlier-platform-based-studies}

An earlier version of this program ran the same readout comparison
inside a commercial simulation platform, with target digests, an
order-reversal replication, a repeated-draws study and a cardinality
manipulation. Those runs motivated the present design and are not
reported here; no claim in this paper depends on them, and every
confirmatory result was re-collected on named models through public
APIs.

\subsection{Appendix G. Coverage
statistics}\label{appendix-g.-coverage-statistics}

For each battery and model, the respondent-by-option matrix is built on
the 0-100 scale of the readout and reduced to three threshold-free
statistics: the number of options whose across-respondent mean is
exactly zero (zero-coverage); the number whose across-respondent
standard deviation is exactly zero (degenerate, which includes options
every respondent selects); and the summed human prevalence of the
zero-coverage options (unexpressed human prevalence). No threshold is
applied anywhere, and only the last requires the human targets.

{\def\LTcaptype{none} 
{\footnotesize\begin{longtable}[]{@{}
  >{\raggedright\arraybackslash}p{(\linewidth - 10\tabcolsep) * \real{0.1667}}
  >{\raggedright\arraybackslash}p{(\linewidth - 10\tabcolsep) * \real{0.1667}}
  >{\raggedright\arraybackslash}p{(\linewidth - 10\tabcolsep) * \real{0.1667}}
  >{\raggedright\arraybackslash}p{(\linewidth - 10\tabcolsep) * \real{0.1667}}
  >{\raggedright\arraybackslash}p{(\linewidth - 10\tabcolsep) * \real{0.1667}}
  >{\raggedright\arraybackslash}p{(\linewidth - 10\tabcolsep) * \real{0.1667}}@{}}
\toprule\noalign{}
\begin{minipage}[b]{\linewidth}\raggedright
Battery
\end{minipage} & \begin{minipage}[b]{\linewidth}\raggedright
Model
\end{minipage} & \begin{minipage}[b]{\linewidth}\raggedright
Options
\end{minipage} & \begin{minipage}[b]{\linewidth}\raggedright
Zero-cov. direct
\end{minipage} & \begin{minipage}[b]{\linewidth}\raggedright
Zero-cov. prob.
\end{minipage} & \begin{minipage}[b]{\linewidth}\raggedright
Unexpressed (pp)
\end{minipage} \\
\midrule\noalign{}
\endhead
\bottomrule\noalign{}
\endlastfoot
CC24\_300 & haiku45 & 5 & 1 & 0 & 3.5 \\
CC24\_300 & gpt4omini & 5 & 1 & 0 & 3.5 \\
CC24\_300 & gemini37flash & 5 & 1 & 0 & 3.5 \\
CC24\_300d & haiku45 & 6 & 4 & 0 & 89.4 \\
CC24\_300d & gpt4omini & 6 & 1 & 0 & 22.2 \\
CC24\_300d & gemini37flash & 6 & 2 & 0 & 36.5 \\
IN5A & haiku45 & 9 & 1 & 0 & 4.0 \\
IN5A & gpt4omini & 9 & 1 & 0 & 4.0 \\
IN5A & gemini37flash & 9 & 1 & 0 & 4.0 \\
OPN & haiku45 & 11 & 4 & 0 & 41.0 \\
OPN & gpt4omini & 11 & 1 & 0 & 14.0 \\
OPN & gemini37flash & 11 & 4 & 0 & 41.0 \\
QA3 & haiku45 & 16 & 10 & 0 & 108.0 \\
QA3 & gpt4omini & 16 & 9 & 0 & 101.0 \\
QA3 & gemini37flash & 16 & 8 & 0 & 73.0 \\
CC24\_305 & claude-haiku-4-5-20251001 & 13 & 9 & 0 & 47.3 \\
CC24\_305 & gpt-4o-mini & 13 & 3 & 0 & 10.1 \\
CC24\_305 & claude-sonnet-5 & 13 & 9 & 0 & 66.7 \\
CC24\_305 & gpt-5.6-terra & 13 & 7 & 0 & 44.5 \\
CC24\_305 & gemini-3.7-flash & 13 & 8 & 0 & 44.6 \\
\end{longtable}}
}

For each zero-coverage row we descriptively compute
\((1-p_H)^{n_{\mathrm{prompt}}}\), the probability of no selections
under calibrated independent draws at the number of distinct prompts.
This is not a valid \(p\)-value for the collected responses: prompt
repetitions are dependent, the synthetic process need not be calibrated,
and uncertainty in the human target is omitted. It is a surprise screen
only. Of the 85 all-instrument zeros, 18 have values at or above 0.05,
33 fall between \(10^{-3}\) and 0.05, and 34 fall below \(10^{-3}\).

The screen is strongly prevalence-dependent: every value at or above
0.05 belongs to an option at or below 4.0 percent human prevalence, and
every value below \(10^{-3}\) belongs to an option at or above 8.1
percent. We do not apply or interpret FDR or Bonferroni procedures
because these quantities are not calibrated hypothesis-test
\(p\)-values. The claim is the observed zero-coverage count itself, with
its sample-size and dependence limitations.
\texttt{analysis/zero\_surprise\_neff.py} reproduces the descriptive
screen.

Pooled, 85 of 206 option slots have zero coverage under the direct
readout and 0 of 206 under the probability readout. The slots are not
independent tests: they are 60 distinct options crossed with three to
five models, and an option that collapses on every model is counted once
per model. By option, 33 of the 60 collapse on at least one model and 12
on every model; of the twelve, four are none or other codes and eight
are substantive (married, had a child and victim of a crime on the
life-events battery; taxation, government debt, the international
situation, security and defence, and Russia's invasion of Ukraine on the
German battery). Degeneracy in the sense of 2.1, an across-respondent
standard deviation of exactly zero, covers those 85 plus a further 15
slots that every respondent selects, so 100 of 206 slots carry no
between-respondent variance at all under the committed-set contract and
none do under the probability contract. The comparison is asymmetric by
construction: a continuous readout avoids zero coverage whenever it
carries a positive floor, so the probability column evidences marginal
expressivity rather than a demonstration that these personas would emit
those options in a realized answer set.

The unexpressed options are not obscure. On the United States
life-events battery they include visiting an emergency room (human 22.1
percent), marrying, divorcing, having a child and being a victim of
crime; on the German issue battery, government debt (12.0) on all three
models, and immigration (23.0) and unemployment (3.0) on two of the
three. The worst single run leaves 108.0 points of summed human
prevalence unexpressed.

\subsection{Appendix H. Scoring detail and
safeguards}\label{appendix-h.-scoring-detail-and-safeguards}

\subsubsection{H.1 Scoring formalism and base
restriction}\label{h.1-scoring-formalism-and-base-restriction}

For battery \(b\) with option set \(\mathcal{O}_b\) and scored base
\(R_b\), let \(S_r\) be respondent \(r\)'s committed set and
\(p_{r,o} \in [0, 100]\) the probability it reports for option \(o\).
The two readouts estimate each option's share as

\[\hat{q}^{\mathrm{set}}_{o} = \frac{100}{|R_b|}\sum_{r \in R_b} \mathbf{1}[o \in S_r], \qquad \hat{q}^{\mathrm{prob}}_{o} = \frac{1}{|R_b|}\sum_{r \in R_b} p_{r,o},\]

and readout \(c\) is scored as
\(\mathrm{MAE}_c(b) = |\mathcal{O}_b|^{-1}\sum_{o} |\hat{q}^{c}_{o} - q_o|\)
against the target shares \(q_o\). Coverage counts exact zeros;
degeneracy counts zero across-response standard deviation.
National-battery intervals resample distinct prompts and are conditional
on the retained cohort, model and collection. CC24\_300d is
sensitivity-only: eligibility comes from the persona's own CC24\_300
committed set (449, 324 and 482 eligible personas), so the evaluation
base is endogenous. Moreover, gpt-4o-mini returns 131 empty arrays among
324 records accepted as valid. Literal-empty scoring gives direct MAE
16.30 and \(\Delta=+4.70\); interpreting empty as the explicit ``None of
the above'' option gives direct MAE 13.16 and \(\Delta=+1.57\).
Excluding empty arrays instead gives direct MAE 24.98 and
\(\Delta=+13.38\). This unresolved coding sensitivity is why no
CC24\_300d number contributes to a headline summary.

\subsubsection{H.2 Safeguards and what they do not
cover}\label{h.2-safeguards-and-what-they-do-not-cover}

Question and option text is retained with the artifact. On the German
battery the spontaneous-only codes ``None'' and ``Don't know'', both at
0 percent, were not shown, while ``Other'' was shown as ``Sonstiges''
and scored against 1 percent. The study notes describe design locks and
partial analyst target blinding, but the release contains no trusted
timestamp or external registry that proves their chronology. SHA-256
digests allow a reader to verify the current bytes against the manifest;
they do not prove when those bytes existed, who had seen them, or what a
model knew. None of the public instruments is model-held-out.

\subsubsection{H.3 Cohort construction
check}\label{h.3-cohort-construction-check}

Allocation audits state that the larger demographic files match supplied
one-way margins and that, because no cohort-safe relationships were
available, variables were assigned independently with a fixed seed. The
runs use their first 500 rows, whose one-way deviations reach 5.4
points. A sensitivity cohort reallocates 500 rows by largest remainder
over the \emph{empirical profile cells of that generated US file}; it
reduces deviation from that file's empirical joint distribution, not
from an official population joint table.

{\def\LTcaptype{none} 
{\footnotesize\begin{longtable}[]{@{}
  >{\raggedright\arraybackslash}p{(\linewidth - 8\tabcolsep) * \real{0.2000}}
  >{\raggedright\arraybackslash}p{(\linewidth - 8\tabcolsep) * \real{0.2000}}
  >{\raggedright\arraybackslash}p{(\linewidth - 8\tabcolsep) * \real{0.2000}}
  >{\raggedright\arraybackslash}p{(\linewidth - 8\tabcolsep) * \real{0.2000}}
  >{\raggedright\arraybackslash}p{(\linewidth - 8\tabcolsep) * \real{0.2000}}@{}}
\toprule\noalign{}
\begin{minipage}[b]{\linewidth}\raggedright
Cohort
\end{minipage} & \begin{minipage}[b]{\linewidth}\raggedright
Direct MAE
\end{minipage} & \begin{minipage}[b]{\linewidth}\raggedright
Probability MAE
\end{minipage} & \begin{minipage}[b]{\linewidth}\raggedright
Difference
\end{minipage} & \begin{minipage}[b]{\linewidth}\raggedright
Zero-coverage direct / probability
\end{minipage} \\
\midrule\noalign{}
\endhead
\bottomrule\noalign{}
\endlastfoot
First 500 rows (used throughout) & 16.80 & 9.55 & +7.26 & 1 / 0 \\
Reallocated over generated-file profile cells & 16.76 & 9.34 & +7.43 & 1
/ 0 \\
\end{longtable}}
}

The paired difference moves by 0.17 points and coverage is unchanged.
This bounds sensitivity to taking the first 500 rows for one battery and
model; it does not establish population-representative joint
composition. The reallocated files ship in
\texttt{cohorts\_stratified/}.

\subsection{Appendix I. Held-out unpublished-cell
targets}\label{appendix-i.-held-out-unpublished-cell-targets}

Every confirmatory target in this paper is a published national table,
so pretraining exposure cannot be excluded by design and Appendix B.4's
probe can only rule out literal recall. This study removes the published
table.

\textbf{Targets.} From the CES 2024 common content microdata we computed
weighted CC24\_305 marginals for four-way demographic cells (gender by
age band by Census region by education). These cell tables are not
published; only the national table is. The retained study record
describes this selection rule: restrict to cells with at least 400
microdata respondents; within each age band take the cell whose
marginals diverge most in mean absolute terms from the national
marginals; then take the three age bands whose best cell diverges most.
Divergence is a property of the human data alone and uses no simulator
output. The study was specified after the confirmatory batteries had
reported. Its record says the cell-selection rule and target file
preceded these calls, and the target file's current SHA-256 begins
e476f436, but the digest does not independently verify that chronology.
The rule deliberately selects high-divergence cells: a model that
recites the national table instead of conditioning on the stated
demographics pays 7.00 points on average. Because the same microdata
both selects the cells and estimates their targets, that 7.00 is subject
to a winner's curse. Re-estimating each selected cell on a random half
of its respondents (seed 20260901; \texttt{heldout/selection\_check.py})
moves the mean divergence from 7.00 to 6.97 points. Scoring the
simulated panels against those selection-free targets leaves the
qualitative result unchanged: the national table scores 6.97 against the
panel's best readout at 7.52 and beats it in three of six cell-model
pairs, against a gap of 0.65 points and the same three of six on the
original targets.

{\def\LTcaptype{none} 
{\footnotesize\begin{longtable}[]{@{}
  >{\raggedright\arraybackslash}p{(\linewidth - 14\tabcolsep) * \real{0.1250}}
  >{\raggedright\arraybackslash}p{(\linewidth - 14\tabcolsep) * \real{0.1250}}
  >{\raggedright\arraybackslash}p{(\linewidth - 14\tabcolsep) * \real{0.1250}}
  >{\raggedright\arraybackslash}p{(\linewidth - 14\tabcolsep) * \real{0.1250}}
  >{\raggedright\arraybackslash}p{(\linewidth - 14\tabcolsep) * \real{0.1250}}
  >{\raggedright\arraybackslash}p{(\linewidth - 14\tabcolsep) * \real{0.1250}}
  >{\raggedright\arraybackslash}p{(\linewidth - 14\tabcolsep) * \real{0.1250}}
  >{\raggedright\arraybackslash}p{(\linewidth - 14\tabcolsep) * \real{0.1250}}@{}}
\toprule\noalign{}
\begin{minipage}[b]{\linewidth}\raggedright
Cell (n microdata)
\end{minipage} & \begin{minipage}[b]{\linewidth}\raggedright
Model
\end{minipage} & \begin{minipage}[b]{\linewidth}\raggedright
Direct
\end{minipage} & \begin{minipage}[b]{\linewidth}\raggedright
Prob.
\end{minipage} & \begin{minipage}[b]{\linewidth}\raggedright
Diff (95\% CI)
\end{minipage} & \begin{minipage}[b]{\linewidth}\raggedright
Zero-cov. dir/prob
\end{minipage} & \begin{minipage}[b]{\linewidth}\raggedright
No-persona query
\end{minipage} & \begin{minipage}[b]{\linewidth}\raggedright
Recite national
\end{minipage} \\
\midrule\noalign{}
\endhead
\bottomrule\noalign{}
\endlastfoot
Man / 18-29 / South / High school or less (460) & haiku45 & 20.56 & 8.51
& +12.05 {[}11.97, 12.11{]} & 11/0 & 9.26 & 8.77 \\
Man / 18-29 / South / High school or less (460) & gpt4omini & 11.27 &
6.32 & +4.95 {[}4.02, 5.91{]} & 6/1 & 8.36 & 8.77 \\
Man / 30-44 / South / High school or less (546) & haiku45 & 18.86 & 8.87
& +9.98 {[}9.60, 10.34{]} & 10/0 & 6.20 & 6.31 \\
Man / 30-44 / South / High school or less (546) & gpt4omini & 8.38 &
11.15 & -2.77 {[}-3.26, -2.29{]} & 6/1 & 9.39 & 6.31 \\
Woman / 65+ / South / 4-year degree (604) & haiku45 & 7.34 & 3.80 &
+3.53 {[}3.19, 3.87{]} & 11/1 & 2.90 & 5.93 \\
Woman / 65+ / South / 4-year degree (604) & gpt4omini & 11.91 & 10.03 &
+1.88 {[}1.74, 1.99{]} & 10/8 & 8.72 & 5.93 \\
\end{longtable}}
}

\textbf{Collection.} Prompts, contracts, decoding and validation are
identical to the base-model replication of Appendix B.1; only the four
stated demographic attributes change. Two models answered as 500
personas per cell under both contracts, 6,000 calls, all valid on the
first or a re-asked attempt. The no-persona prevalence query of Appendix
B.3 was additionally asked about each cell's population, twenty
repetitions per cell per model.

{\def\LTcaptype{none} 
{\footnotesize\begin{longtable}[]{@{}
  >{\raggedright\arraybackslash}p{(\linewidth - 10\tabcolsep) * \real{0.1667}}
  >{\raggedright\arraybackslash}p{(\linewidth - 10\tabcolsep) * \real{0.1667}}
  >{\raggedright\arraybackslash}p{(\linewidth - 10\tabcolsep) * \real{0.1667}}
  >{\raggedright\arraybackslash}p{(\linewidth - 10\tabcolsep) * \real{0.1667}}
  >{\raggedright\arraybackslash}p{(\linewidth - 10\tabcolsep) * \real{0.1667}}
  >{\raggedright\arraybackslash}p{(\linewidth - 10\tabcolsep) * \real{0.1667}}@{}}
\toprule\noalign{}
\begin{minipage}[b]{\linewidth}\raggedright
Cell
\end{minipage} & \begin{minipage}[b]{\linewidth}\raggedright
Model
\end{minipage} & \begin{minipage}[b]{\linewidth}\raggedright
Direct
\end{minipage} & \begin{minipage}[b]{\linewidth}\raggedright
Probability
\end{minipage} & \begin{minipage}[b]{\linewidth}\raggedright
Difference
\end{minipage} & \begin{minipage}[b]{\linewidth}\raggedright
Original difference
\end{minipage} \\
\midrule\noalign{}
\endhead
\bottomrule\noalign{}
\endlastfoot
Man / 18-29 / South / High school or less & haiku45 & 20.59 & 7.01 &
+13.58 & +12.05 \\
Man / 18-29 / South / High school or less & gpt4omini & 10.67 & 6.71 &
+3.96 & +4.95 \\
Man / 30-44 / South / High school or less & haiku45 & 14.91 & 5.41 &
+9.50 & +9.98 \\
Man / 30-44 / South / High school or less & gpt4omini & 8.51 & 8.76 &
-0.25 & -2.77 \\
Woman / 65+ / South / 4-year degree & haiku45 & 5.89 & 5.00 & +0.89 &
+3.53 \\
Woman / 65+ / South / 4-year degree & gpt4omini & 11.54 & 11.17 & +0.37
& +1.88 \\
\end{longtable}}
}

\textbf{Readings.} The readout effect survives on targets that cannot
have been memorised: five of six comparisons favour probabilities, from
+1.88 to +12.05, every interval excluding zero. Support collapse is
worse here than on national targets, at 6 to 11 zero-coverage options of
13. Against that, gpt-4o-mini on the 30-44 cell reverses, with the
direct readout better by 2.77 {[}-3.26, -2.29{]}; this is the only
reversal observed outside the capped battery, and the account below
traces it to the rare-option regime of Appendix A.2. And the probability
readout is not structurally immune to support collapse: on the 65+ cell
gpt-4o-mini returns exactly zero on 8 of 13 options, which is direct
evidence for the asymmetry caveat of Section 2.1 rather than against it,
since the contract's clean coverage elsewhere is an empirical property
and not a guarantee.

\textbf{Wording control.} This study removes the published target but
not the published item wording, so we re-ran all six conditions with the
question and every option label paraphrased: semantically matched,
lexically distinct (``visited an emergency room'' becomes ``went to a
hospital emergency department'', and so on; the full mapping is in
heldout/paraphrase.py). The human target belongs to the original
wording, so this tests whether the readout effect depends on the
published phrasing, not whether the paraphrased instrument measures the
same construct.

Five of six still favour probabilities, the range is -0.25 to +13.58
against the original -2.77 to +12.05, and the per-condition effect sizes
correlate at r = 0.94. Neither the target nor the phrasing being the
published one leaves the effect intact.

\textbf{The reversal, explained.} The gpt-4o-mini result on the 30-44
cell is the one negative comparison, and it follows from the mechanism
of Appendix A.2 rather than contradicting it. That cell's human
prevalences are unusually low: total selected mass 145.2 percent against
the national 157.7, with ``none of the above'' at 36.2 percent against a
national 19.3. The direct readout under-selects to a total mass of 50.0
while the probability readout over-selects to 198.9, and because most of
this cell's options are rare, the direct contract's collapse is cheap
while the probability contract's positive floor is expensive: married
(human 1.9) draws 0.0 direct against 22.2 probability, had a child
(human 1.9) 0.4 against 17.7. This is the rare-tercile regime in which
Appendix A.2 already reports the direct contract ahead by 4.3 points.
Across the six conditions the cell's mean human option share correlates
with the readout advantage at \(r = 0.47\), directional but weak on six
points and confounded with model, so we rest the account on the
within-cell decomposition rather than on that correlation.

\textbf{Cost.} The five-model replication of Appendix B.1 cost 4.63 US
dollars in provider charges; its archived cost fields use the pricing
assumptions documented in the artifact. Token-usage records for the
other API runs are included in the release package.

\subsection{Appendix J. Subgroup conditioning at thirty
cells}\label{appendix-j.-subgroup-conditioning-at-thirty-cells}

The held-out study of Appendix I uses three cells and reports that
neither contract beats reciting the national table. Three cells cannot
distinguish absent subgroup signal from signal swamped by level bias, so
we expanded the descriptive analysis. Cells with at least 300 microdata
respondents were sorted by divergence from the national table and
sampled at even rank intervals across 3.07 to 8.77 points, rather than
only from the tail. This reduces the specific tail-selection issue but
does not make the cells a random sample. Two hundred persona calls per
cell and contract were run on claude-haiku-4-5-20251001 and
gemini-3.7-flash; 12,000 records were valid on the former and 11,975 on
the latter. The retained record says targets preceded collection, but
its digest does not prove chronology. Because divergence-rank sampling
widens the human variation, cell-to-cell correlations below are
descriptive upper bounds.

{\def\LTcaptype{none} 
{\footnotesize\begin{longtable}[]{@{}
  >{\raggedright\arraybackslash}p{(\linewidth - 12\tabcolsep) * \real{0.1429}}
  >{\raggedright\arraybackslash}p{(\linewidth - 12\tabcolsep) * \real{0.1429}}
  >{\raggedright\arraybackslash}p{(\linewidth - 12\tabcolsep) * \real{0.1429}}
  >{\raggedright\arraybackslash}p{(\linewidth - 12\tabcolsep) * \real{0.1429}}
  >{\raggedright\arraybackslash}p{(\linewidth - 12\tabcolsep) * \real{0.1429}}
  >{\raggedright\arraybackslash}p{(\linewidth - 12\tabcolsep) * \real{0.1429}}
  >{\raggedright\arraybackslash}p{(\linewidth - 12\tabcolsep) * \real{0.1429}}@{}}
\toprule\noalign{}
\begin{minipage}[b]{\linewidth}\raggedright
Model
\end{minipage} & \begin{minipage}[b]{\linewidth}\raggedright
Contract
\end{minipage} & \begin{minipage}[b]{\linewidth}\raggedright
MAE vs cell truth
\end{minipage} & \begin{minipage}[b]{\linewidth}\raggedright
Reciting the national table
\end{minipage} & \begin{minipage}[b]{\linewidth}\raggedright
Per-option bias
\end{minipage} & \begin{minipage}[b]{\linewidth}\raggedright
Cell-to-cell correlation
\end{minipage} & \begin{minipage}[b]{\linewidth}\raggedright
Per-option median \(r\)
\end{minipage} \\
\midrule\noalign{}
\endhead
\bottomrule\noalign{}
\endlastfoot
claude-haiku-4-5 & Committed set & 11.36 & 4.85 & 8.94 & \(+0.394\) &
\(+0.473\) (4 of 13 options) \\
claude-haiku-4-5 & Probability & 6.08 & 4.85 & 4.62 & \(+0.543\) &
\(+0.787\) (13 of 13) \\
gemini-3.7-flash & Committed set & 12.14 & 4.85 & 10.69 & \(+0.505\) &
\(+0.688\) (5 of 13) \\
gemini-3.7-flash & Probability & 5.09 & 4.85 & 4.48 & \(+0.826\) &
\(+0.855\) (13 of 13) \\
\end{longtable}}
}

Per-option bias is the mean over options of the absolute value of that
option's mean signed error across cells, that is, the part of the error
a constant per-option offset would remove.
\texttt{conditioning/score\_conditioning.py} recomputes every entry.

At these thirty selected cells, neither contract has lower MAE than the
national-table baseline. A post hoc decomposition asks whether level
bias masks cell ordering. After centering each option across cells,
simulated and human deviations correlate at \(r = 0.54\) pooled with a
per-option median of \(0.79\) for one probability model; the second
model gives \(r = +0.826\) and a median of \(+0.855\). These
correlations may reflect demographic conditioning, but the rank-spread
cell selection inflates available variation and the analysis has no
survey-statistical comparator. The committed-set column is also
difficult to compare: only 4 of 13 options vary across cells for one
model, so its median is computed on that restricted set. We therefore
treat the decomposition as a hypothesis about biased levels, not as
evidence that the panel reliably recovers subgroup structure.

This suggests that demographic conditioning is not information-free, but
that its levels are poorly calibrated in these selected cells. The
decomposition was specified after the three-cell result and the
correlations are conditional on a rank-spread sample, so it is
hypothesis-generating. Appendix J.1 explores an offset correction on the
same data.

\subsubsection{J.1 Does anchoring the levels repair
it?}\label{j.1-does-anchoring-the-levels-repair-it}

The decomposition motivates a descriptive correction. If the subgroup
error is mostly a per-option level offset, estimating that offset on
some cells may reduce error on the remaining cells in the same pool. We
evaluate that calculation on the same runs, with no further model calls.
For a budget of \(k\) anchor cells, we draw \(k\) cells at random,
estimate one offset per option as the mean signed error on those cells,
apply it to every other cell, clip to \([0, 100]\) and score only the
held-out cells; anchors are never scored. Two hundred draws per budget,
seed 20260901; the anchor draws are shared across the two models, so the
national-table columns are identical for both.

We also apply the offset procedure to the national table so that both
calculations consume the same human cell data. Differences between the
two are descriptive within-pool contrasts; they do not isolate a causal
contribution from simulated ordering or establish performance on new
cells or batteries.

{\def\LTcaptype{none} 
{\footnotesize\begin{longtable}[]{@{}
  >{\raggedright\arraybackslash}p{(\linewidth - 10\tabcolsep) * \real{0.1667}}
  >{\raggedright\arraybackslash}p{(\linewidth - 10\tabcolsep) * \real{0.1667}}
  >{\raggedright\arraybackslash}p{(\linewidth - 10\tabcolsep) * \real{0.1667}}
  >{\raggedright\arraybackslash}p{(\linewidth - 10\tabcolsep) * \real{0.1667}}
  >{\raggedright\arraybackslash}p{(\linewidth - 10\tabcolsep) * \real{0.1667}}
  >{\raggedright\arraybackslash}p{(\linewidth - 10\tabcolsep) * \real{0.1667}}@{}}
\toprule\noalign{}
\begin{minipage}[b]{\linewidth}\raggedright
Anchor cells
\end{minipage} & \begin{minipage}[b]{\linewidth}\raggedright
Probability, anchored
\end{minipage} & \begin{minipage}[b]{\linewidth}\raggedright
Committed set, anchored
\end{minipage} & \begin{minipage}[b]{\linewidth}\raggedright
National table
\end{minipage} & \begin{minipage}[b]{\linewidth}\raggedright
National, anchored the same way
\end{minipage} & \begin{minipage}[b]{\linewidth}\raggedright
Draws where anchored probability beats the national table
\end{minipage} \\
\midrule\noalign{}
\endhead
\bottomrule\noalign{}
\endlastfoot
\emph{claude-haiku-4-5} & & & & & \\
none & 6.08 & 11.36 & 4.85 & --- & --- \\
1 & 5.62 & 8.32 & 4.85 & 6.49 & 15.0\% \\
2 & 4.96 & 7.89 & 4.85 & 5.59 & 45.5\% \\
3 & 4.76 & 7.71 & 4.86 & 5.34 & 65.0\% \\
5 & 4.55 & 7.53 & 4.84 & 5.06 & 86.5\% \\
10 & 4.42 & 7.45 & 4.87 & 4.98 & 95.0\% \\
20 & 4.36 & 7.28 & 4.85 & 4.86 & 90.0\% \\
\emph{gemini-3.7-flash} & & & & & \\
none & 5.09 & 12.14 & 4.85 & --- & --- \\
1 & 3.52 & 7.91 & 4.85 & 6.49 & 96.0\% \\
2 & 3.20 & 7.45 & 4.85 & 5.59 & 100.0\% \\
3 & 3.01 & 7.32 & 4.86 & 5.34 & 100.0\% \\
5 & 2.90 & 7.12 & 4.84 & 5.06 & 100.0\% \\
10 & 2.77 & 6.99 & 4.87 & 4.98 & 100.0\% \\
20 & 2.73 & 6.88 & 4.85 & 4.86 & 100.0\% \\
\end{longtable}}
}

Descriptively, ten anchors take claude-haiku-4-5 probabilities from 6.08
to 4.42, with 95.0 percent of the 200 held-cell draws below the
unadjusted national table; one anchor takes gemini-3.7-flash from 5.09
to 3.52, with 96.0 percent below it. These rates are conditional
resampling frequencies, not probabilities of success on a new battery.
Committed sets remain above the national table at all displayed budgets.
Applying the same offsets to the cell-invariant table increases its
error because it has no ordering to preserve. That control helps
interpret the calculation, but it is not a fitted small-area or
multilevel survey estimator.

This analysis does not license a general anchor budget or superiority
claim. It shows that, on two models and one selected CES battery,
per-option offsets estimated on randomly chosen cells reduce held-cell
MAE. The correction and evaluation reuse the same thirty-cell pool and
were specified post hoc. \texttt{conditioning/anchor\_repair.py}
reproduces the table.

\subsubsection{J.2 Replication on a second
battery}\label{j.2-replication-on-a-second-battery}

The thirty-cell design was repeated on CES CC24\_300 (media use, five
options), the other CES battery with microdata, after Appendix J had
reported. Thirty cells were selected by the same rule (at least 300
microdata respondents, sorted by divergence from the national table,
taken at even rank intervals; 78 candidates, divergence 2.5 to 15.1
points). The retained record says the target preceded calls;
\texttt{target\_digest\_cc24\_300.txt} authenticates its current bytes
but not that chronology. Two hundred personas per cell under both
contracts, with prompts and decoding as in Appendix J, on
claude-haiku-4-5, claude-sonnet-5, gpt-5.6-terra and gemini-3.7-flash,
four models from three providers, 48,000 valid responses of 48,000
requested. claude-sonnet-5 and gpt-5.6-terra run at their default
temperatures where the API refuses the requested one, as in Appendix
B.1.

{\def\LTcaptype{none} 
{\footnotesize\begin{longtable}[]{@{}
  >{\raggedright\arraybackslash}p{(\linewidth - 12\tabcolsep) * \real{0.1429}}
  >{\raggedright\arraybackslash}p{(\linewidth - 12\tabcolsep) * \real{0.1429}}
  >{\raggedright\arraybackslash}p{(\linewidth - 12\tabcolsep) * \real{0.1429}}
  >{\raggedright\arraybackslash}p{(\linewidth - 12\tabcolsep) * \real{0.1429}}
  >{\raggedright\arraybackslash}p{(\linewidth - 12\tabcolsep) * \real{0.1429}}
  >{\raggedright\arraybackslash}p{(\linewidth - 12\tabcolsep) * \real{0.1429}}
  >{\raggedright\arraybackslash}p{(\linewidth - 12\tabcolsep) * \real{0.1429}}@{}}
\toprule\noalign{}
\begin{minipage}[b]{\linewidth}\raggedright
Model
\end{minipage} & \begin{minipage}[b]{\linewidth}\raggedright
Contract
\end{minipage} & \begin{minipage}[b]{\linewidth}\raggedright
MAE vs cell truth
\end{minipage} & \begin{minipage}[b]{\linewidth}\raggedright
Reciting the national table
\end{minipage} & \begin{minipage}[b]{\linewidth}\raggedright
Per-option bias
\end{minipage} & \begin{minipage}[b]{\linewidth}\raggedright
Cell-to-cell correlation
\end{minipage} & \begin{minipage}[b]{\linewidth}\raggedright
Per-option median \(r\)
\end{minipage} \\
\midrule\noalign{}
\endhead
\bottomrule\noalign{}
\endlastfoot
claude-haiku-4-5 & Committed set & 28.56 & 7.57 & 17.89 & \(+0.353\) &
\(+0.485\) (4 of 5) \\
claude-haiku-4-5 & Probability & 11.18 & 7.57 & 10.67 & \(+0.760\) &
\(+0.723\) (5 of 5) \\
claude-sonnet-5 & Committed set & 25.98 & 7.57 & 15.06 & \(+0.471\) &
\(+0.468\) (4 of 5) \\
claude-sonnet-5 & Probability & 8.21 & 7.57 & 6.83 & \(+0.862\) &
\(+0.802\) (5 of 5) \\
gpt-5.6-terra & Committed set & 22.49 & 7.57 & 15.50 & \(+0.711\) &
\(+0.735\) (4 of 5) \\
gpt-5.6-terra & Probability & 6.67 & 7.57 & 5.18 & \(+0.854\) &
\(+0.879\) (5 of 5) \\
gemini-3.7-flash & Committed set & 24.46 & 7.57 & 12.75 & \(+0.677\) &
\(+0.659\) (4 of 5) \\
gemini-3.7-flash & Probability & 6.33 & 7.57 & 4.33 & \(+0.880\) &
\(+0.860\) (5 of 5) \\
\end{longtable}}
}

The same descriptive pattern appears on an instrument with different
content and five rather than thirteen options. Unanchored probability
MAE is 11.18, 8.21, 6.67 and 6.33 against the table's 7.57, with pooled
cell-order correlations from \(r=0.76\) to \(0.88\) and substantial
level bias. Two models already outperform the table before offsets.
Under committed sets, one of five options returns an identical share in
every cell on every model.

{\def\LTcaptype{none} 
{\footnotesize\begin{longtable}[]{@{}
  >{\raggedright\arraybackslash}p{(\linewidth - 10\tabcolsep) * \real{0.1667}}
  >{\raggedright\arraybackslash}p{(\linewidth - 10\tabcolsep) * \real{0.1667}}
  >{\raggedright\arraybackslash}p{(\linewidth - 10\tabcolsep) * \real{0.1667}}
  >{\raggedright\arraybackslash}p{(\linewidth - 10\tabcolsep) * \real{0.1667}}
  >{\raggedright\arraybackslash}p{(\linewidth - 10\tabcolsep) * \real{0.1667}}
  >{\raggedright\arraybackslash}p{(\linewidth - 10\tabcolsep) * \real{0.1667}}@{}}
\toprule\noalign{}
\begin{minipage}[b]{\linewidth}\raggedright
Anchor cells
\end{minipage} & \begin{minipage}[b]{\linewidth}\raggedright
Probability, anchored
\end{minipage} & \begin{minipage}[b]{\linewidth}\raggedright
Committed set, anchored
\end{minipage} & \begin{minipage}[b]{\linewidth}\raggedright
National table
\end{minipage} & \begin{minipage}[b]{\linewidth}\raggedright
National, anchored the same way
\end{minipage} & \begin{minipage}[b]{\linewidth}\raggedright
Draws where anchored probability beats the national table
\end{minipage} \\
\midrule\noalign{}
\endhead
\bottomrule\noalign{}
\endlastfoot
\emph{claude-haiku-4-5} & & & & & \\
1 & 8.09 & 16.98 & 7.58 & 10.35 & 39.0\% \\
2 & 7.34 & 17.13 & 7.58 & 9.06 & 69.0\% \\
3 & 6.85 & 17.75 & 7.56 & 8.66 & 87.5\% \\
5 & 6.55 & 17.36 & 7.59 & 8.23 & 94.5\% \\
10 & 6.34 & 17.23 & 7.52 & 7.86 & 99.0\% \\
20 & 6.23 & 17.45 & 7.59 & 7.83 & 94.5\% \\
\emph{claude-sonnet-5} & & & & & \\
1 & 6.29 & 16.69 & 7.58 & 10.35 & 85.5\% \\
2 & 5.74 & 17.09 & 7.58 & 9.06 & 97.0\% \\
3 & 5.41 & 17.07 & 7.56 & 8.66 & 99.5\% \\
5 & 5.19 & 17.15 & 7.59 & 8.23 & 100.0\% \\
10 & 5.04 & 16.83 & 7.52 & 7.86 & 100.0\% \\
20 & 4.88 & 16.72 & 7.59 & 7.83 & 100.0\% \\
\emph{gpt-5.6-terra} & & & & & \\
1 & 6.95 & 12.41 & 7.58 & 10.35 & 72.5\% \\
2 & 6.43 & 12.78 & 7.58 & 9.06 & 89.0\% \\
3 & 6.02 & 12.68 & 7.56 & 8.66 & 97.0\% \\
5 & 5.77 & 12.68 & 7.59 & 8.23 & 99.5\% \\
10 & 5.58 & 12.66 & 7.52 & 7.86 & 100.0\% \\
20 & 5.44 & 12.27 & 7.59 & 7.83 & 99.5\% \\
\emph{gemini-3.7-flash} & & & & & \\
1 & 6.69 & 16.54 & 7.58 & 10.35 & 76.5\% \\
2 & 6.10 & 17.32 & 7.58 & 9.06 & 93.5\% \\
3 & 5.71 & 17.18 & 7.56 & 8.66 & 98.5\% \\
5 & 5.43 & 17.38 & 7.59 & 8.23 & 100.0\% \\
10 & 5.19 & 17.09 & 7.52 & 7.86 & 100.0\% \\
20 & 5.09 & 16.85 & 7.59 & 7.83 & 100.0\% \\
\end{longtable}}
}

Within these 200-draw summaries, the first budget reaching a 95-percent
draw-win rate is 10, 2, 3 and 3 cells across the four models. These are
descriptive thresholds on selected cells, not prospective sample-size
recommendations. Committed sets remain above the table at every
displayed budget. \texttt{conditioning/score\_cc24\_300.sh} reproduces
both tables.

\subsubsection{J.3 Richer grounding on the thirty
cells}\label{j.3-richer-grounding-on-the-thirty-cells}

The objection to Appendix J is that four attributes are the weakest
grounding anyone would use. We therefore reran the thirty CC24\_305
cells on claude-haiku-4-5 with the elaboration most persona products
perform: for each of the 200 personas per cell the model first writes a
short first-person profile from the cell's four attributes (work,
household, daily life), and the battery is then answered under both
contracts conditioned on that profile rather than on the bare
attributes. The profile adds no information about the person beyond the
four attributes; it changes how they are presented. 6,000 profiles were
written and 12,000 of 12,000 answers were valid.
\texttt{conditioning/runner\_conditioning\_profile.py} collects the arm
and the scorers of Appendix J score it with
\texttt{COND\_TAG=profile\_\_}.

{\def\LTcaptype{none} 
{\footnotesize\begin{longtable}[]{@{}
  >{\raggedright\arraybackslash}p{(\linewidth - 10\tabcolsep) * \real{0.1667}}
  >{\raggedright\arraybackslash}p{(\linewidth - 10\tabcolsep) * \real{0.1667}}
  >{\raggedright\arraybackslash}p{(\linewidth - 10\tabcolsep) * \real{0.1667}}
  >{\raggedright\arraybackslash}p{(\linewidth - 10\tabcolsep) * \real{0.1667}}
  >{\raggedright\arraybackslash}p{(\linewidth - 10\tabcolsep) * \real{0.1667}}
  >{\raggedright\arraybackslash}p{(\linewidth - 10\tabcolsep) * \real{0.1667}}@{}}
\toprule\noalign{}
\begin{minipage}[b]{\linewidth}\raggedright
Grounding
\end{minipage} & \begin{minipage}[b]{\linewidth}\raggedright
Contract
\end{minipage} & \begin{minipage}[b]{\linewidth}\raggedright
MAE vs cell truth
\end{minipage} & \begin{minipage}[b]{\linewidth}\raggedright
Per-option bias
\end{minipage} & \begin{minipage}[b]{\linewidth}\raggedright
Cell-to-cell correlation
\end{minipage} & \begin{minipage}[b]{\linewidth}\raggedright
Per-option median \(r\)
\end{minipage} \\
\midrule\noalign{}
\endhead
\bottomrule\noalign{}
\endlastfoot
Four attributes & Committed set & 11.36 & 8.94 & \(+0.394\) & \(+0.473\)
(4 of 13) \\
Four attributes & Probability & 6.08 & 4.62 & \(+0.543\) & \(+0.787\)
(13 of 13) \\
Written profile & Committed set & 9.63 & 9.20 & \(+0.358\) & \(+0.351\)
(9 of 13) \\
Written profile & Probability & 4.52 & 3.39 & \(+0.784\) & \(+0.702\)
(13 of 13) \\
\end{longtable}}
}

Reciting the national table scores 4.85 on the same cells. The profile
helps: under the probability contract the per-option bias moves from
4.62 to 3.39, the MAE from 6.08 to 4.52, below the table, and the pooled
ordering correlation from \(r = 0.54\) to \(0.78\), while the per-option
median falls from \(0.79\) to \(0.70\). The committed-set contract moves
from 11.36 to 9.63, with 9 of 13 options now varying across cells
against 4 before, and the readout effect persists at \(+5.11\) against
\(+5.29\). Anchoring the profile arm as in J.1 (bare / profile):

{\def\LTcaptype{none} 
{\footnotesize\begin{longtable}[]{@{}
  >{\raggedright\arraybackslash}p{(\linewidth - 8\tabcolsep) * \real{0.2000}}
  >{\raggedright\arraybackslash}p{(\linewidth - 8\tabcolsep) * \real{0.2000}}
  >{\raggedright\arraybackslash}p{(\linewidth - 8\tabcolsep) * \real{0.2000}}
  >{\raggedright\arraybackslash}p{(\linewidth - 8\tabcolsep) * \real{0.2000}}
  >{\raggedright\arraybackslash}p{(\linewidth - 8\tabcolsep) * \real{0.2000}}@{}}
\toprule\noalign{}
\begin{minipage}[b]{\linewidth}\raggedright
Anchor cells
\end{minipage} & \begin{minipage}[b]{\linewidth}\raggedright
Probability, anchored
\end{minipage} & \begin{minipage}[b]{\linewidth}\raggedright
Committed set, anchored
\end{minipage} & \begin{minipage}[b]{\linewidth}\raggedright
National table
\end{minipage} & \begin{minipage}[b]{\linewidth}\raggedright
Draws beating the national table
\end{minipage} \\
\midrule\noalign{}
\endhead
\bottomrule\noalign{}
\endlastfoot
1 & 5.62 / 4.15 & 8.32 / 6.12 & 4.85 & 15.0\% / 81.0\% \\
2 & 4.96 / 3.67 & 7.89 / 5.67 & 4.85 & 45.5\% / 98.0\% \\
3 & 4.76 / 3.47 & 7.71 / 5.45 & 4.86 & 65.0\% / 100.0\% \\
5 & 4.55 / 3.32 & 7.53 / 5.22 & 4.84 & 86.5\% / 100.0\% \\
10 & 4.42 / 3.19 & 7.45 / 5.09 & 4.87 & 95.0\% / 100.0\% \\
\end{longtable}}
}

In this post hoc arm, the first budget reaching a 95-percent empirical
draw-win rate changes from 10 cells with bare attributes to 2 with
generated profiles. Narrative elaboration is therefore not inert on this
battery and model, but it adds no individual evidence and does not
establish a transferable anchor budget. A level bias of 3.39 remains,
and committed sets remain collapsed. Grounding on real individual
evidence remains untested.

\subsection{Appendix K.
Reproducibility}\label{appendix-k.-reproducibility}

Every number in this paper is produced by a script in the release
package from retained response records and target files; no commercial
simulation platform contributes to a confirmatory result. The package
includes the six battery targets and their SHA-256 digests in
\texttt{target\_digests.json} and \texttt{MANIFEST.sha256}. Rehashing
verifies the bytes shipped in this archive, but neither a digest nor the
accompanying study note independently establishes when those bytes were
fixed or who had inspected them.

For model calls, the retained JSONL generally contains the terminal
model text, parsed answer, validity flag, run-local respondent
identifier, requested model alias and collection parameters. It is not a
complete archive of provider transport envelopes, and token usage is
run-level for some collections and absent for others; exact per-call
routing and billing provenance should therefore not be inferred. The
package covers the battery-model comparisons (\texttt{allbatteries/},
\texttt{base\_model\_replication/}), temperature and retest collections
(\texttt{base\_model\_replication/}, \texttt{retest/},
\texttt{matched\_temp/}), cohort sensitivity
(\texttt{allbatteries\_strat/}, \texttt{cohorts\_stratified/}),
no-persona queries and probes (\texttt{base\_model\_replication/}),
held-out cells (\texttt{heldout/}), and exploratory conditioning
analyses (\texttt{conditioning/}). \texttt{release/README.md} maps
tables to scripts. \texttt{scripts/verify\_numbers.py} checks manuscript
values against derived files; \texttt{scripts/audit\_numbers.py}
independently rebuilds statistics from retained responses and targets
where possible and checks released intermediate files otherwise.

The CES 2024 common content microdata is public but not redistributed;
scripts that require it read \texttt{CES\_MICRODATA}. No human free
text, persona names or platform identifiers are included; Appendix J.3
profiles are model-generated and respondent identifiers are run-local.

\end{document}